\documentclass[10pt,twocolumn,letterpaper]{article}

\usepackage{wacv}
\usepackage{times}
\usepackage{epsfig}
\usepackage{graphicx}
\usepackage{amsfonts}
\usepackage[ruled]{algorithm2e}
\usepackage{amsmath}
\usepackage{amssymb}
\usepackage{multicol}
\usepackage{multirow}
\usepackage{diagbox}
\usepackage{array}
\usepackage{siunitx}
\usepackage{float}

\wacvfinalcopy 

\def\wacvPaperID{0794} 

\ifwacvfinal\fi
\begin{document}

\title{Image denoising via K-SVD with primal-dual active set algorithm}


\author{Quan Xiao \hspace{2.5cm} Canhong Wen* \hspace{2.5cm} Zirui Yan\\
Department of Statistics and Finance\\School of Management\\University of Science and Technology of China\\
{\tt\small xq2016@mail.ustc.edu.cn \hspace{0.6cm} wench@ustc.edu.cn \hspace{0.6cm} zincrain@mail.ustc.edu.cn}
}

\maketitle
\ifwacvfinal\thispagestyle{empty}\fi

\begin{abstract}

   K-SVD algorithm has been successfully applied to image denoising tasks dozens of years but the big bottleneck in speed and accuracy still needs attention to break. For the sparse coding stage in K-SVD, which involves $\ell_{0}$ constraint, prevailing methods usually seek approximate solutions greedily but are less effective once the noise level is high. The alternative $\ell_{1}$ optimization is proved to be powerful than $\ell_{0}$, however, the time consumption prevents it from the implementation. In this paper, we propose a new K-SVD framework called K-SVD$_P$ by applying the Primal-dual active set (PDAS) algorithm to it. Different from the greedy algorithms based K-SVD, the K-SVD$_P$ algorithm develops a selection strategy motivated by KKT (Karush-Kuhn-Tucker) condition and yields to an efficient update in the sparse coding stage. Since the K-SVD$_P$ algorithm seeks for an equivalent solution to the dual problem iteratively with simple explicit expression in this denoising problem, speed and quality of denoising can be reached simultaneously. Experiments are carried out and demonstrate the comparable denoising performance of our K-SVD$_P$ with state-of-the-art methods.

\end{abstract}

\let\thefootnote\relax\footnotetext{*Canhong Wen is the corresponding author.}

\section{Introduction}
Image denoising problem is primal in various regions such as image processing and computer visions. The goal of denoising is to remove noise from noisy images and retain the actual signal as precisely as possible. Many methods based on sparse representation have been proposed to accomplish this goal in the past few decades \cite{Claus05,Alpher25,Alpher26,Alpher27,Alpher29,baloch2017image}. K-means singular value decomposition (K-SVD) is one of the typical works among these models. It is an iterative patch-based procedure aiming at finding an optimal linear combination of an overcomplete dictionary to best describe the image. The solid theoretical foundations \cite{Alpher33} and adaptability make it boost for dozens of years. It can be divided into two stages, one is the dictionary learning stage and the other is the sparse coding stage. Some recent researches have been seeking for highly efficient ways to make a breakthrough, but these modifications mostly are taken on the dictionary learning stage \cite{Alpher30,Alpher31}.

In fact, sparse coding is an optimization problem and $\ell_{1}$ optimization \cite{Alpher15,Alpher20,Alpher18} is proved more powerful in solving denoising problems when the noise level is high \cite{hastie2017extended}. However, taking time consumption into consideration, the image denoising area always perfers to approximate the $\ell_{0}$ solutions using greedy algorithms instead \cite{Alpher23} and treats it as benchmark of K-SVD \cite{Alpher07,Alpher23,Claus05,Alpher33,Claus06,abdelhamed2018high}. Recently, Liu \etal \cite{liu2019mixed}apply the Mixed Integer quadratic programming (MIQP) in the sparse coding stage which yields the global optimal solution, but it also takes a long time. Thus, a tradeoff between computational efficiency and denoising performance in high noise conditions is needed. 

In this paper, primal-dual active set algorithm (PDAS) is applied to the sparse coding stage in the K-SVD framework, and the new framework is called K-SVD$_P$. PDAS algorithm is first proposed by Ito and Kunisch in 2013 \cite{Alpher35}and then generalized and implemented by Wen, Zhang \etal in 2017 \cite{aawen2017bess}. By using the KKT condition and introducing the primal-dual variables, this NP-hard problem\cite{Alpher37} can be switched to a restricted linear regression model which can be solved explicitly. We demonstrate the feasibility of this new scheme and compare it with the existing K-SVD models achieved by orthogonal matching pursuit (OMP) algorithm\cite{tropp2004greed,tropp2007signal}, a typical $\ell_{0}$ greedy algorithm, and Basis pursuit Denoising(BPDN) algorithm (also known as LASSO in statistics) \cite{Alpher15,sajjad2016basis}, a classic $\ell_{1}$ optimization algorithm, in experiment. The potential of our method will be verified both theoretically and experimentally.

These are our major contributions:
\begin{itemize}
\item We successfully build a new K-SVD$_P$ framework by applying the PDAS algorithm to sparse coding stage and reach an explicit expression in this special case;
\item Comparison with the representative algorithms OMP and BPDN are taken both theoretically and empirically;
\item The results demonstrate the proposed K-SVD$_P$ is competitive when the noise is low and has superior performance in highly noisy images compared to the-state-of-art methods.
\end{itemize}

The rest of the paper is organized as follows. In Section 2, we state the image denoising problem and introduce the K-SVD framework. In Section 3, K-SVD$_P$ is proposed and theoretical analysis is described. In Section 4, experiments in image denoising are carried out and the results are showed. In Section 5, we arrive at the conclusion and mention the possible future work.

\section{Problem statement and K-SVD framework}
Image denoising problem can be described as $\beta=\alpha+\epsilon$, where $\alpha$ is the original noise-free image, $\epsilon$ is the added random Gaussian white noise, and $\beta$ denotes the noisy image. Our target is to move $\epsilon$ from given $\beta$ and obtain the real image $\alpha$.

In order to achieve this, sparse representation model first searches for the principal component of the image called dictionary by extracting sparse elements patch by patch in $\beta$, and then treats the residual as noise $\epsilon$ and throw it out, and finally reconstruct the image $\alpha$ based on the sparse representation of the selected image elements. In this paper, we only focus on the first phase of the above procedure which the K-SVD algorithm is designed for and the other details can be found in \cite{Alpher23}.

\begin{figure}
\begin{tabular}{ccccccccc}
\fbox{\includegraphics[width=2cm]{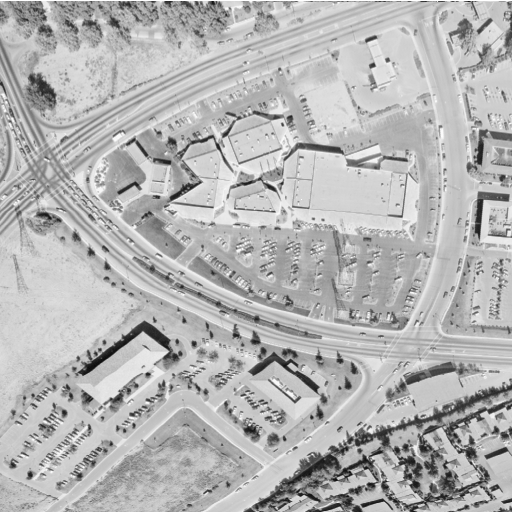}}&
\fbox{\includegraphics[width=2cm]{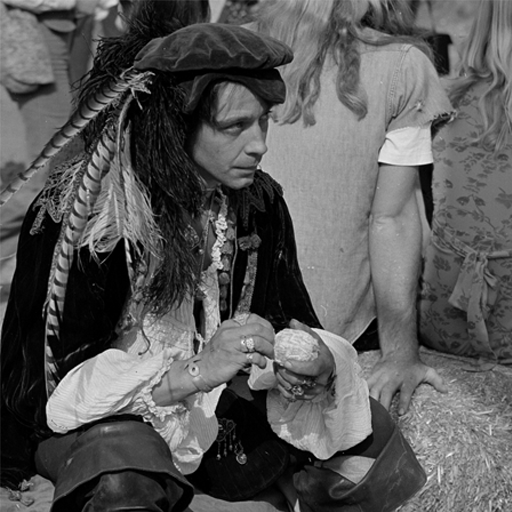}}&
\fbox{\includegraphics[width=2cm]{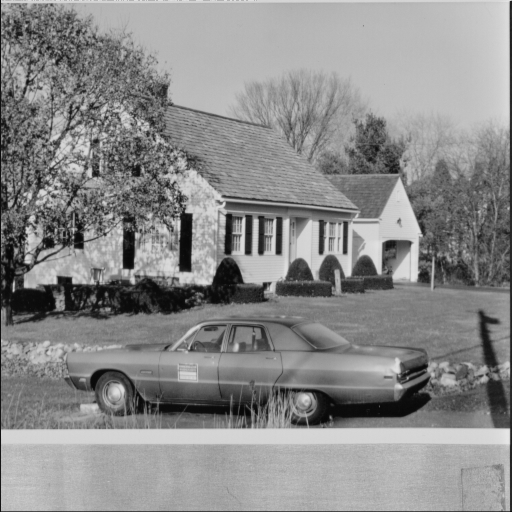}}&\\
\footnotesize(a) Map&\footnotesize(b) Man&\footnotesize(c) House&\\
\fbox{\includegraphics[width=2cm]{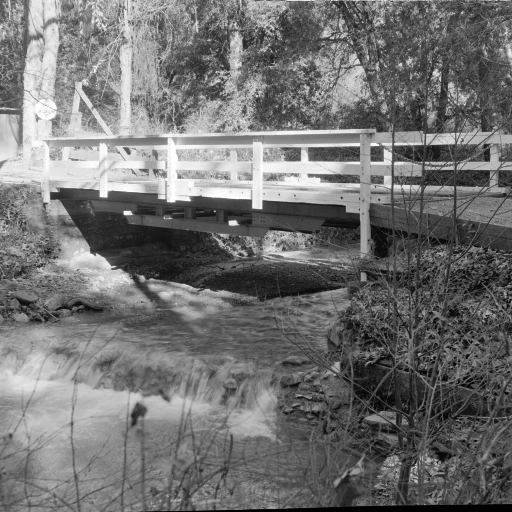}}&
\fbox{\includegraphics[width=2cm]{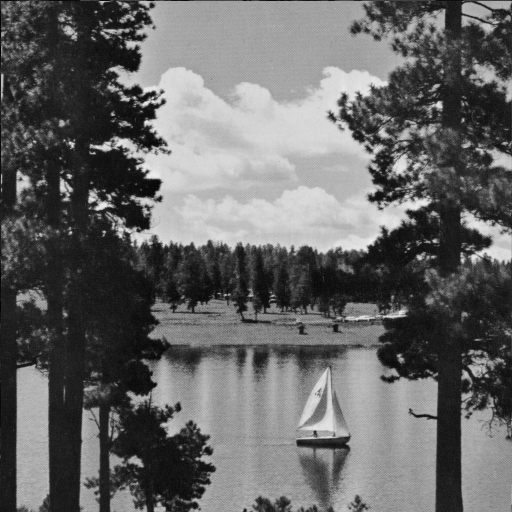}}&
\fbox{\includegraphics[width=2cm]{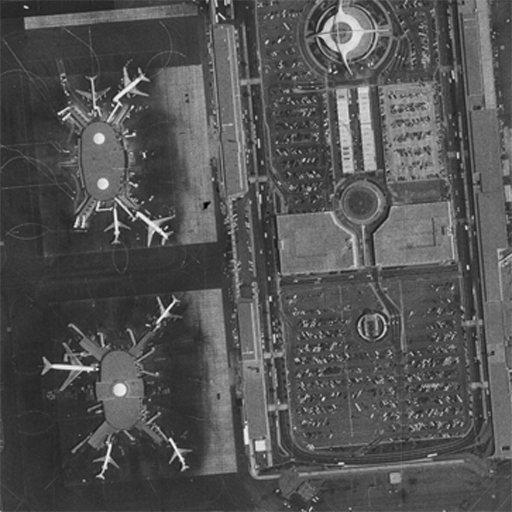}}\\
\footnotesize(d) Bridge&\footnotesize (e) Lake&\footnotesize(f) Airport&\\
\fbox{\includegraphics[width=2cm]{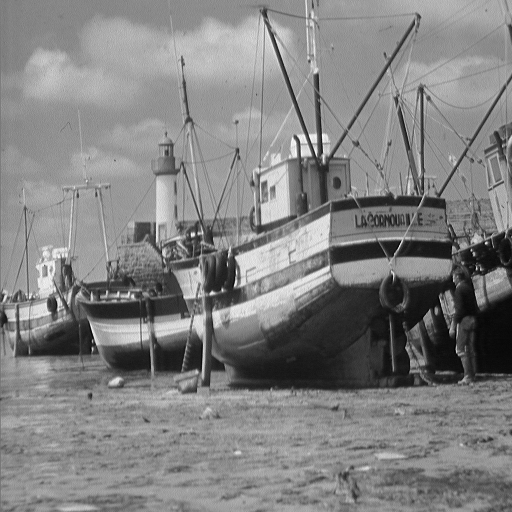}}&
\fbox{\includegraphics[width=2cm]{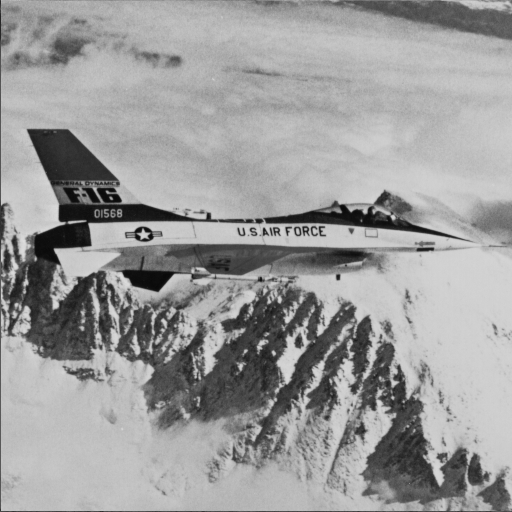}}&
\fbox{\includegraphics[width=2cm]{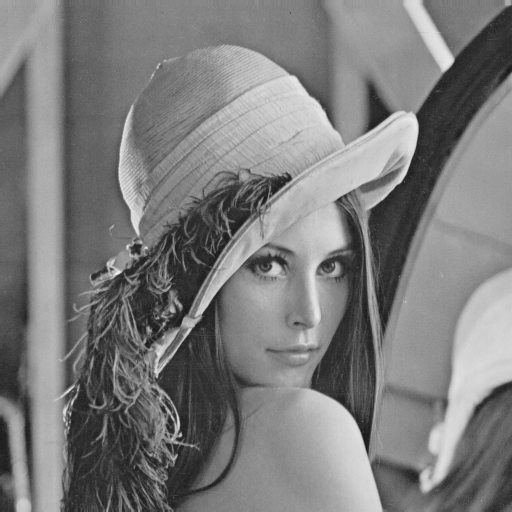}}&\\
\footnotesize(g) Boat&\footnotesize(h) Airplane&\footnotesize(i) Lena&
\end{tabular}
\\
\caption{Chosen images from USC-SIPI Image Database}
\label{fig:chosen}
\end{figure}

Considering a signal matrix $Y=\{y_{j}\}_{j=1}^{p} \in \mathbb{R}^{n \times p}$ with $p$ original signals, a dictionary $D=\{d_{j}\}_{j=1}^{K} \in \mathbb{R}^{n \times K}$ with $K$ prototype signal-atoms and sparse representation $X=\{x_{j}\}_{j=1}^{p} \in \mathbb{R}^{K \times p}$ with $p$ solutions $x_{j}$ of corresponding $y_{j}$. The optimization object can be formulated as:
\begin{equation}
\arg \min _{D, X}\left\{\|Y-D X\|_{F}^{2}\right\} \quad s.t. \left\|x_{i}\right\|_{0} \leq T_{0}, i=1,2, \cdots, p
\end{equation}
where $T_{0}$ is the sparsity level, \ie $\ell_{0}$-$norm$ counting the number of nonzero elements in a vector, and $\|Y-DX\|_{F}^{2}=\sum\left\|y_{i}-Dx_{i}\right\|_{2}^{2}$, \ie the the Frobenius Norm of matrix $Y-DX$.

K-SVD algorithm consists of dictionary learning and sparse coding stage. The dictionary learning stage is to update the dictionary and corresponding coefficient with given $X$, and the sparse coding stage deals with finding the sparse coefficient $x_{i}$ to each $y_{i}$ with known dictionary $D$. To simplify the formula at sparse coding stage, let $y$ and $x$ denote $y_{i}$ and $x_{i}$, the target is as follows:
\begin{equation}
\hat{x}=\arg \min \left\|y-D x\right\|_{2}^{2} \quad s.t. \left\|x\right\|_{0} \leqslant T_{0}.
\end{equation}

The dictionary learning stage is generally solved by applying Single-Value Decomposition (SVD) to nonzero submatrix of each ${E}_{i}=Y-\sum_{j \neq i} {d}_{j} {x}_{(j)}$, where ${x}_{(j)}$ denotes the $j-th$ row of $X$ since the first column of singular value vector contains the highest proportion of information. That is, to extract the first column of the left singular value vector to update atoms column and treat the first column of the right singular value vector as the corresponding coefficient column. The details can be found in \cite{Alpher07}. 

While the dictionary updating stage generates a convex optimization problem, the sparse coding stage with $\ell_{0}$-$norm$ constraint is more challenging. 

\section{Proposed pursuit algorithm}
Since the problem in (2) is NP-hard problem\cite{Alpher37}, prevailing algorithms usually search for approximate solutions by greedy algorithms (\eg Matching pursuit \cite{bergeaud1995matching}, Orthogonal matching pursuit \cite{tropp2004greed}). However, most of these approaches suffer from insufficient precision in high noise level \cite{jhang2016high}. A remedy is turning to solve $\ell_{1}$ optimization (\eg Basis pursuit\cite{chen2001atomic}, Basis pursuit Denoising \cite{Alpher15,sajjad2016basis}) which has promising accuracy equivalently \cite{hastie2017extended}, but the computational expense makes it infeasible in large-scale denoising problems. This really obstructs the development of the K-SVD framework in image denoising, since at least thousands of patches are in the process even if for the small $124\times124$ image.

In this section, we plug a special case of the PDAS algorithm, proposed by Wen \etal \cite{aawen2017bess} who derived KKT condition for general convex loss functions, in the K-SVD$_P$ sparse coding stage. The goal of this section is to derive an explicit expression in the denoising problem, and then discuss the connection with existing approaches.

\subsection{The K-SVD$_P$ sparse coding stage}
It's known that solution to (2) is necessarily a coordinate-wise minimizer. So, let $x^{\diamond}=(x_{1}^{\diamond}, \dots, x_{K}^{\diamond})$ be the coordinate-wise minimizer, \ie each $x_{j}^{\diamond}$ is minimizer in its coordinate. A simple observation is that:
\begin{equation}
\begin{array}{l l}
\left\|y-D x\right\|_{2}^{2} &=\sum\limits_{i=1}^{n} (y_{i}-\sum\limits_{q=1}^{K} D_{iq}x_{q})^{2}\\
&=\sum\limits_{i=1}^{n} (y_{i}-\sum\limits_{q\neq j} D_{iq}x_{q}-D_{ij}x_{j})^{2}\\
&=\sum\limits_{i=1}^{n} (y_{i}-\sum\limits_{q\neq j} D_{iq}x_{q})^{2}\\
&-2\sum\limits_{i=1}^{n}(y_{i}-\sum\limits_{q\neq j} D_{iq}x_{q})D_{ij}x_{j}+\sum\limits_{i=1}^{n} {D_{ij}}^{2}{x_{j}}^{2}\\
&=\sum\limits_{i=1}^{n} (y_{i}-\sum\limits_{q\neq j} D_{iq}x_{q})^{2}\\
&-2\sum\limits_{i=1}^{n}(y_{i}-\sum\limits_{q\neq j} D_{iq}x_{q})D_{ij}x_{j}+{x_{j}}^{2}
\end{array}
\end{equation}
where last equation is arrived since dictionary $D$ is normalized.

In order to find coordinate-wise minimizer, we define a quadratic function respective to $t$ in each coordinate $j$ which freezes $x$ in the other coordinates to their optimal choices:
\begin{equation}
l_{j}(t)=\sum_{i=1}^{n} (y_{i}-\sum_{q\neq j}D_{iq}x_{q}^{\diamond})^{2}-2\sum_{i=1}^{n}(y_{i}-\sum_{q\neq j} D_{iq}x_{q}^{\diamond})D_{ij}t+t^{2}
\end{equation}
Then $l_{j}(t)$ achieve global minimum if and only if $t_j^*=\sum_{i=1}^{n}(y_{i}-\sum_{q\neq j} D_{iq}x_{q}^{\diamond})D_{ij}$. For simplicity, let $d_{j}$ denotes the $j-th$ column of $D$, and define $g_j^\diamond=(y-Dx^\diamond)^{\diamond\prime} d_{j}$. In this way, $t_j^*=x_{j}^\diamond+g_{j}^\diamond$. It's natural to define a sacrifice of $l_{j}(t)$ if we push $t_j^*$ from desirable value $x_{j}^\diamond+g_{j}^\diamond$ to zero, and that is:
\begin{equation}
h_{j}=\frac{1}{2}{(x_j^\diamond+g_j^\diamond)}^2
\end{equation}
We tend to set those scarify less to zero. \ie
\begin{equation}
 x_j^\diamond  = \left\{
  \begin{array}{l l}
    x_j^\diamond + g_j^\diamond , & \text{ if }  h_{j}\geq h_{[T_{0}]}  \\
    0, & \text{ else},
  \end{array} \right. \quad\mbox{for } j=1,\ldots,K,
\end{equation}
Actually, these are the KKT conditions of $x^\diamond$ proved in \cite{aaawen2017bess}. So $x^\diamond$ is the solution to (2) if and only if it satisfies the above conditions. We can tell from (6) that if $x_{j}\neq 0$, then $x_{j}$ is the optimal value and $g_{j}=0$, and if not, $g_{j}\neq 0$ as defined. This observation indicates $x_{j}$ and $g_{j}$ have complementary supports and we can treat them as a pair of primal-dual variables. Then, searching for a solution to (2) is equal to finding the best dual variable $g_{j}$. Let $\mathcal{A}$ be the indicator set of nonzero elements in coefficient $x$ and $\mathcal{I}=\left(\mathcal{A}\right)^{c}$. Then we arrive at:
\begin{equation}
 \left\{
  \begin{array}{l l}
    x_{\mathcal{I}}=0,\\
    g_{\mathcal{A}}=0,\\
    x_{\mathcal{A}}=\left(D_{A}^{\prime} D_{A}\right)^{-1} D_{A}^{\prime} y,\\
    g_{\mathcal{I}}=(y-Dx)^{T} D_{\mathcal{I}},\\
    h_{\mathcal{I}}=\frac{1}{2}{(g_{\mathcal{I}})}^2,\\
    h_{\mathcal{A}}=\frac{1}{2}{(x_{\mathcal{A}})}^2
  \end{array} \right.
\end{equation}
and
\begin{equation}
\mathcal{A}=\left\{j : h_{j} \geq h_{[k]}\right\} ,\quad \mathcal{I}=\left\{j : h_{j}<h_{[k]}\right\}
\end{equation}
where $h_{[1]} \geq h_{[2]} \geq \ldots, \geq h_{[K]}$ denotes the decreasing permutation of $h$. We solve this problem iteratively and reach the pursuit algorithm in Algorithm 1.\\

\begin{algorithm}
            \caption{Sparse coding algorithm in K-SVD$_P$}
            \KwIn{Signal $y$, fixed dictionary $D \in \mathbb{R}^{n \times K}$, the maximum number of iterations $R$ and $\ell_{0}-norm$ constraint $T_{0}$}
            \KwOut{Sparse representation $x$}

            Initialization: randomly set $\mathcal{A}^{0}$ be a $T_{0}$ subset of $\{1, \ldots, K\}$ and $\mathcal{I}^{0}=\left(\mathcal{A}^{0}\right)^{c}$\;
            \For{$r \in\{0,1 \ldots, R\}$}{
                    $\bullet$ Compute $x_{\mathcal{I}}^r,x_{\mathcal{A}}^r,g_{\mathcal{I}}^r,g_{\mathcal{A}}^r,h_{\mathcal{I}}^r,h_{\mathcal{A}}^r$ by equation (6) where $\mathcal{A}=\mathcal{A}^r$\;
                    $\bullet$ Sort $h_{j}$ by $h_{[1]}^{r} \geq h_{[2]}^{r} \geq \ldots, \geq h_{[K]}^{r}$ \;
                    $\bullet$ Update the active and inactive sets by\\

                    \begin{center}
                    $\mathcal{A}^{r+1}=\left\{j : h_{j}^{r} \geq h_{[T_{0}]}^{r}\right\}$,\quad $\mathcal{I}^{r+1}=\left\{j : h_{j}^{r}<h_{[T_{0}]}^{r}\right\}$
                    \end{center}
                    $\bullet$ If $\mathcal{A}^{r+1}=\mathcal{A}^{r}$, then stop; else $r=r+1$ and return to steps above.
        }
\end{algorithm}

\begin{figure}
\begin{center}
\begin{tabular}{cccccc}
\includegraphics[width=3.8cm]{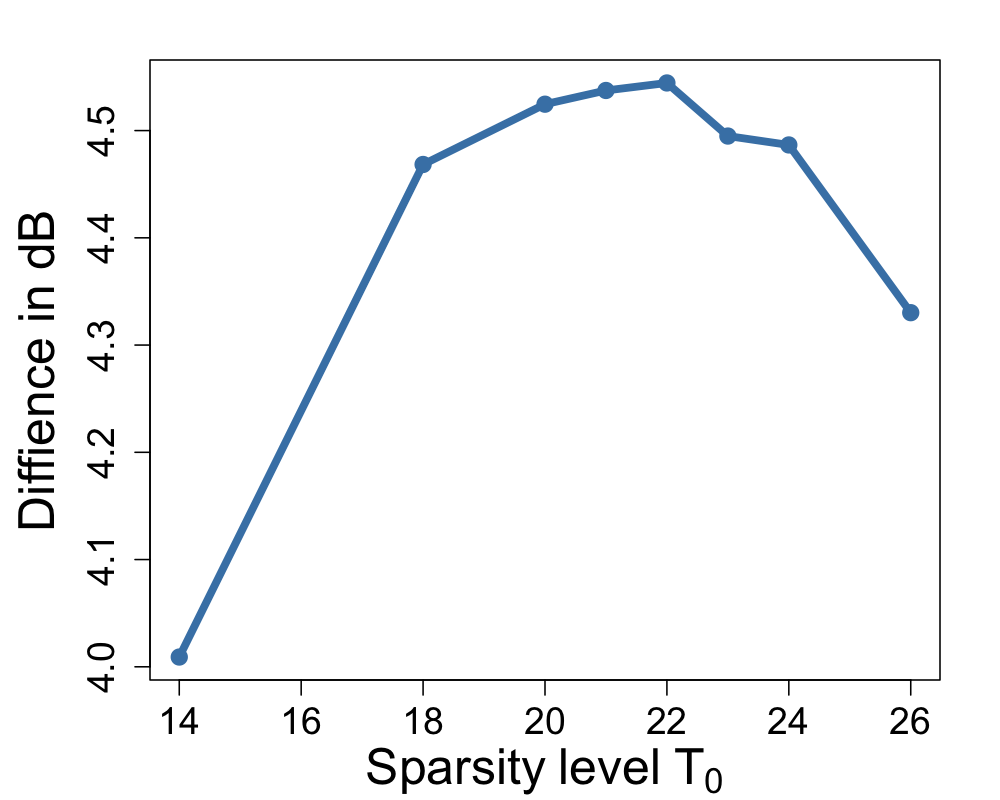}&
\includegraphics[width=3.8cm]{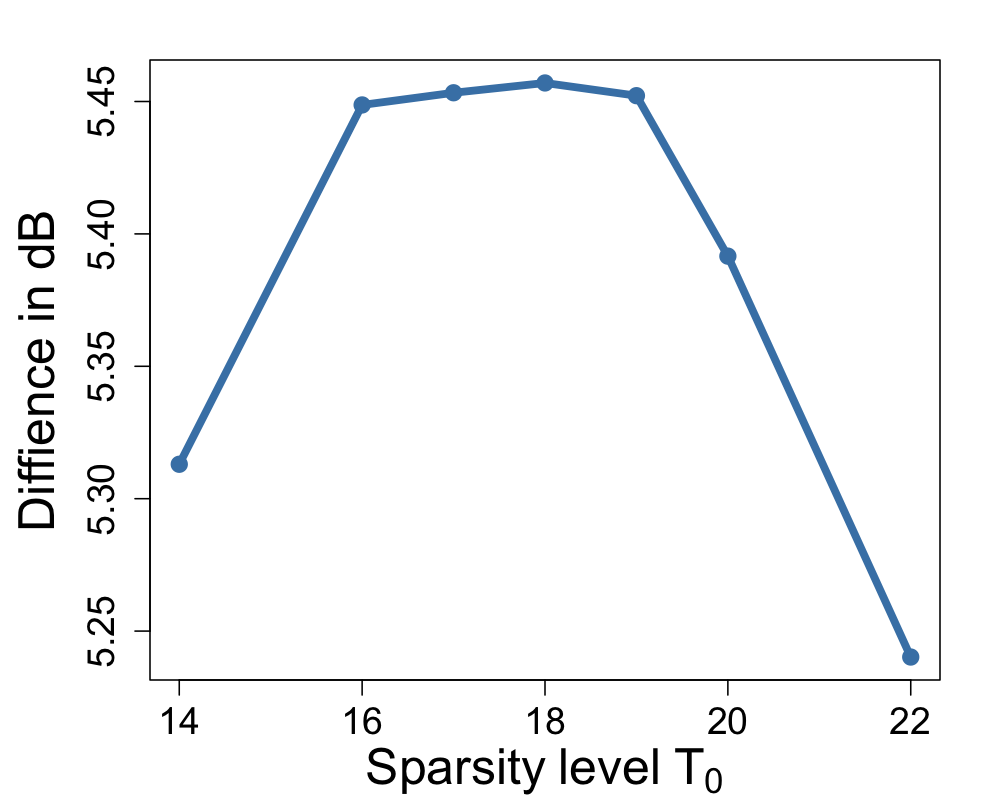}&\\
\footnotesize(a) $\sigma=15$&\footnotesize(b) $\sigma=20$&\\
\includegraphics[width=3.8cm]{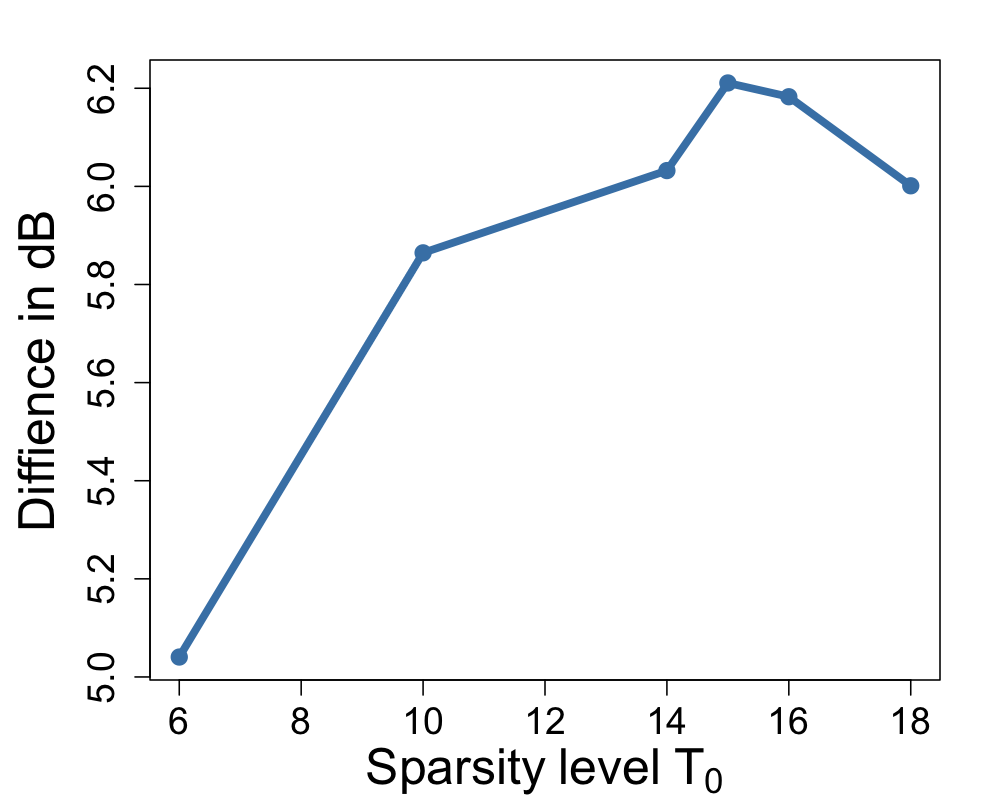}&
\includegraphics[width=3.8cm]{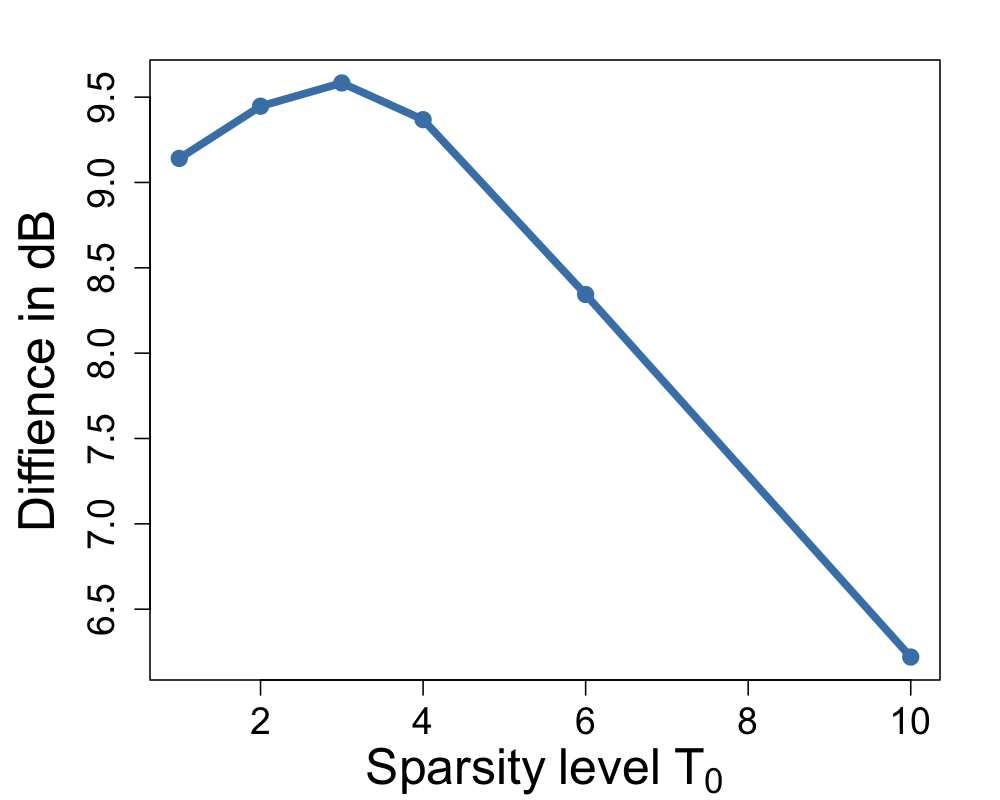}&\\
\footnotesize(c) $\sigma=25$&\footnotesize(d) $\sigma=50$&\\
\includegraphics[width=3.8cm]{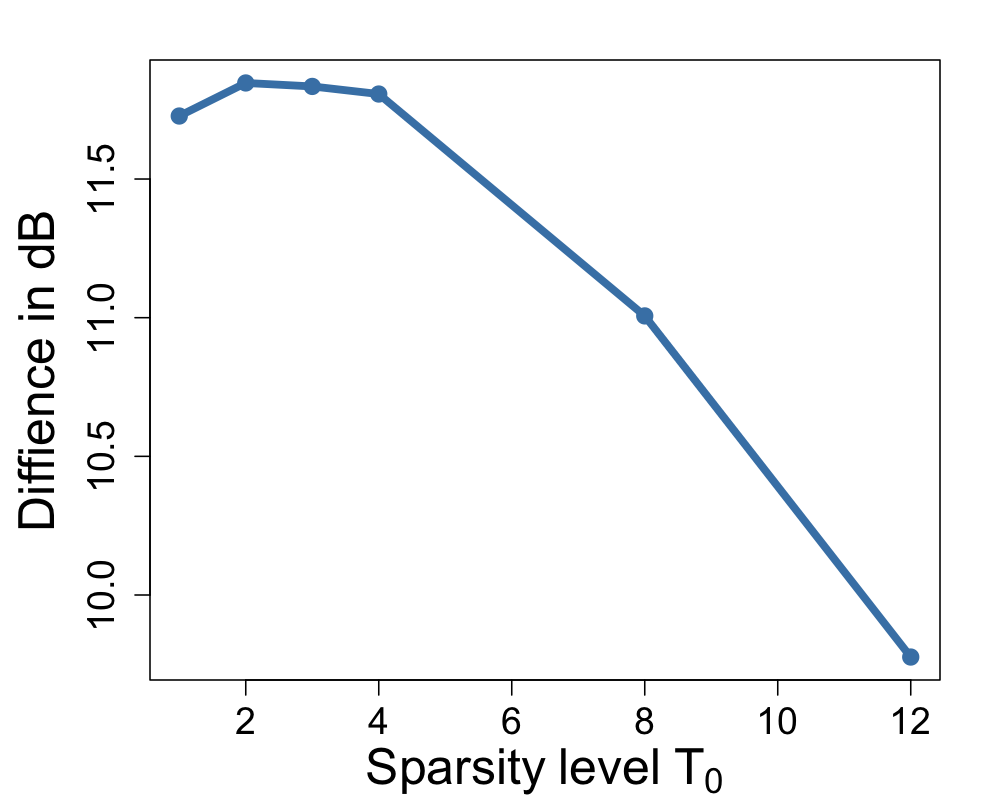}&
\includegraphics[width=3.8cm]{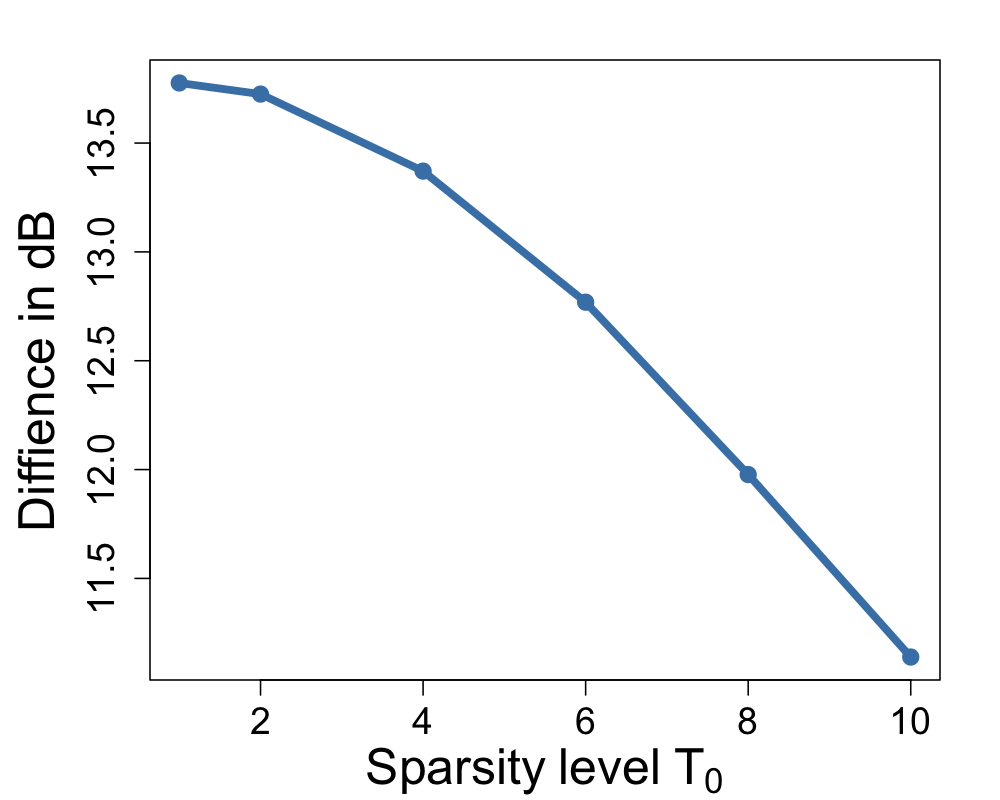}\\
\footnotesize(e) $\sigma=75$&\footnotesize(f) $\sigma=100$&
\end{tabular}
\\
\caption{PSNR of K-SVD$_P$ versus sparsity levels $T_0$ with $\sigma=15,20,25,50,75,100$}
\label{fig:chosen2}
\end{center}
\end{figure}

\subsection{Comparison with existing approach}
\subsubsection{Comparison to greedy algorithms}
In this part, a theoretical comparison to the representative of greedy algorithm Orthogonal matching pursuit algorithm (OMP) \cite{tropp2004greed} is given. OMP algorithm is an iterative method. Let $P_{i}$ be the indicator set of dictionary atoms have been selected until $i$-th step and $R_{i}$ be the residual in $i$-th step, \ie $R_{i}=y-Dx^{(i)}$ where $x^{(i)}$ denotes the sparse coefficient $x$ in $i$-th step. At ($i$+1)-th step, one atom $d_{j}$ that is most correlated to the residual $R_{i}$ is selected by maximizing $|\langle d_{j},R_{i}\rangle|$ which is same as dual variables $g_j=(y-Dx)^{\prime} d_{j}$ defined in the K-SVD$_P$ algorithm. Then, in order to keep new residual orthogonal to selected dictionary atoms, the OMP algorithm estimates the nonzero elements of $x^{(i+1)}$ and computes residual $R_{i+1}$ by applying least squares to $y$ on dictionary atoms have been selected already, \ie $x^{(i+1)}=(D_{P_{i}}^{\prime} D_{P_{i}})^{-1} D_{P_{i}}^{\prime} y$. After several iterations, this algorithm will converge.

However, from equation (5), we can tell that the K-SVD$_P$ algorithm updates $T_{0}$ atoms in the active set each time based on $h_{j}$ which collects information both in primal and dual variables iteratively. This procedure will gather more information in each step, which accelerates the convergence and improves the denoising performance.

\subsubsection{Comparison with $\ell_{1}$ denoising}
Since $\ell_{1}$-$norm$ is the closest convex function to $\ell_{0}$-$norm$, alternative methods seek for $l_{1}$ constraint solution of problem (1). By transforming the $\ell_{0}$-$norm$ into $\ell_{1}$-$norm$ and formulating the Lagrangian function, the problem is changed to basis pursuit denoising(BPDN) problem, which is also known as LASSO \cite{Alpher15} in statistics.
\begin{equation}
\hat{x}(\lambda)=\arg \min \left\|y-D x\right\|_{2}^{2}+\lambda  \left\|x\right\|_{1}
\end{equation}
Recently, Hastie \etal (2017)showed that neither best subset selection (2) nor LASSO (9) dominates in terms of accuracy, with best subset selection performs better in low noisy conditions and LASSO better in highly noisy images\cite{hastie2017extended}. This tolerance of high noise partly thanks to the shrinkage in LASSO, since the fitted variables from LASSO are continuous functions of $x$.\cite{Alpher20} Best subset selection will hit discontinuous points when $x$ is moving from $\mathcal{I}$ to $\mathcal{A}$ or from $\mathcal{A}$ to $\mathcal{I}$  which makes them susceptible to high noise. However, the K-SVD$_P$ based on best subset selection is still attractive since its time complexity is far less than LASSO as shown in the next section. As is said in \cite{tropp2004greed}, if there is an approximant holding of good quality, there is no need to waste time in finding another closer solution.

\section{Experiment}
\subsection{Design and Parameter setting}
We select 9 images of size $512\times 512$, as shown in Figure 1, from classic USC-SIPI Image Database\cite{olshausen1997sparse} to compare the image denoising performance of the K-SVD$_P$ with the OMP and the BPDN-based K-SVD scheme.
\begin{table}
\begin{center}
\begin{tabular}{cccccccc}
\hline
$\sigma$& 15& 20& 25& 50& 75& 100\\
\hline
$T_0$& 20& 20& 15& 2& 2& 2\\
\hline
\end{tabular}
\end{center}
\caption{Chosen sparsity level $T_0$ in $\sigma=15,20,25,50,75,100$}
\end{table}

For similarity, we set the number of iteration of K-SVD to $10$ for all pursuit algorithms. For each image, $p=500$ overlapping patches of size $n=8\times 8$ are extracted to learn the dictionary $D$ of size $64\times 256$ as suggested ~\cite{Alpher23}. The experiment is repeated for noise levels $\sigma=15,20,25,50,75,100$. Note that the last three are the high noise level benchmarks according to ~\cite{Alpher23}. In order to select the optimal sparsity level $T_0$ at different noise levels for K-SVD$_P$, we start with the noisy Man image for the experiment. In each sparsity level for each $\sigma$, we compute the peak signal-to-noise ratio (PSNR) of the restored image. The PSNR of two images $x$ and $y$ is defined as (10). The results are presented in Figure 2. Based on the results, the optimal sparsity levels are chosen in Table 1. 
\begin{equation}
\mathrm{PSNR}=-10 \log \frac{\|x-y\|^{2}}{255^{2}}
\end{equation}

\begin{table}[htbp]
\begin{center}
\resizebox{80mm}{8mm}{
\begin{tabular}{c|ccc|ccc}
\hline
{$\sigma$}& 15& 20& 25& 50& 75& 100\\
\hline
BPDN& 1178.91& 1177.59& 1180.31& -& -& -\\

OMP&       78.04& 80.82& 83.20& 78.76& 79.33&78.37\\

K-SVD$_P$& 93.29& 96.19& 84.28& 58.93& 59.47& 61.56\\
\hline
\end{tabular}}
\end{center}
\caption{Reconstruction time(s) with $\sigma=15,20,25,50,75,100$}
\end{table}

\begin{table*}[!htbp]
\footnotesize
\begin{center}
\resizebox{\textwidth}{37mm}{
\begin{tabular}{c|ccc|ccc|ccc}
\hline
{Figure}& \multicolumn{3}{c|}{Map}& \multicolumn{3}{c|}{Man}& \multicolumn{3}{c}{House}\\
\hline
{$\sigma/PSNR$}&BPDN&OMP&K-SVD$_P$&BPDN&OMP&K-SVD$_P$&BPDN&OMP&K-SVD$_P$\\
\hline
{$15/24.61$}&27.10&28.39&27.83&28.87&29.31&29.04&28.67&29.10&28.71\\
{$20/22.11$}&25.99&26.63&26.38&27.55&27.24&27.50&27.39&27.12&27.08\\
{$25/20.17$}&24.97&25.07&24.98&26.37&25.57&26.09&26.23&25.46&25.88\\
\hline
{Figure}& \multicolumn{3}{c|}{Bridge}& \multicolumn{3}{c|}{Lake}& \multicolumn{3}{c}{Airport}\\
\hline
{$\sigma/PSNR$}&BPDN&OMP&K-SVD$_P$&BPDN&OMP&K-SVD$_P$&BPDN&OMP&K-SVD$_P$\\
\hline
{$15/24.61$}&26.72&27.82&27.34&28.74&29.04&28.71&29.14&29.55&29.41\\
{$20/22.11$}&25.81&26.26&26.15&27.40&27.05&27.15&27.64&27.32&27.58\\
{$25/20.17$}&24.95&24.87&24.99&26.28&25.45&26.14&26.45&25.62&26.48\\
\hline
{Figure}& \multicolumn{3}{c|}{Boat}& \multicolumn{3}{c|}{Airplane}& \multicolumn{3}{c}{Lena}\\
\hline
{$\sigma/PSNR$}&BPDN&OMP&K-SVD$_P$&BPDN&OMP&K-SVD$_P$&BPDN&OMP&K-SVD$_P$\\
\hline
{$15/24.61$}&29.25&29.28&29.12&29.73&29.59&29.39&30.89&29.89&30.07\\
{$20/22.11$}&27.77&27.27&27.42&28.27&27.42&27.52&29.09&27.64&28.27\\
{$25/20.17$}&26.64&25.57&26.42&26.87&25.65&26.62&27.62&25.83&27.45\\
\hline
\end{tabular}
}
\end{center}
\caption{Accuracy of the reconstruction PSNR(in dB) with $\sigma=15,20,25$ (The higher, the better)}
\end{table*}

\begin{table*}[!htbp]
\footnotesize
\begin{center}
\resizebox{\textwidth}{24mm}{
\begin{tabular}{c|cc|cc|cc|cc|cc}
\hline
{Figure}& \multicolumn{2}{c|}{Map}& \multicolumn{2}{c|}{Man}& \multicolumn{2}{c|}{House}&\multicolumn{2}{c|}{Birdge}&\multicolumn{2}{c}{Lake}\\
\hline
{$\sigma/PSNR$}&OMP&K-SVD$_P$&OMP&K-SVD$_P$&OMP&K-SVD$_P$&OMP&K-SVD$_P$&OMP&K-SVD$_P$\\
\hline
{$50/14.15$}&19.75&\bf21.76&19.92&\bf23.74&19.87&\bf23.26&19.68&\bf22.26&19.88&\bf23.35\\
{$75/10.63$}&16.40&\bf20.67&16.48&\bf22.30&16.46&\bf21.85&16.39&\bf21.29&16.47&\bf21.87\\
{$100/8.13$}&13.97&\bf19.83&14.02&\bf21.17&14.00&\bf20.78&13.99&\bf20.26&14.04&\bf20.81\\
\hline
{Figure}& \multicolumn{2}{c|}{Airport}&\multicolumn{2}{c|}{Boat}&\multicolumn{2}{c|}{Airplane}&\multicolumn{2}{c|}{Lena}&\multicolumn{2}{c}{Average}\\
\hline
{$\sigma/PSNR$}&OMP&K-SVD$_P$&OMP&K-SVD$_P$&OMP&K-SVD$_P$&OMP&K-SVD$_P$&OMP&K-SVD$_P$\\
\hline
{$50/14.15$}&19.90&\bf23.85&19.96&\bf24.13&19.96&\bf24.03&19.99&\bf25.67&19.88&\bf23.56\\
{$75/10.63$}&16.48&\bf22.48&16.48&\bf22.61&16.49&\bf22.26&16.54&\bf23.87&16.47&\bf22.13\\
{$100/8.13$}&14.03&\bf21.34&14.02&\bf21.36&14.04&\bf21.11&14.06&\bf22.41&14.02&\bf21.01\\
\hline
\end{tabular}}
\end{center}
\caption{Accuracy of the reconstruction in terms of the PSNR(in dB) with $\sigma=50,75,100$ (The best are highlighted in bold)}
\end{table*}
For the OMP algorithm, we run the supplementary code provided by \cite{Alpher23} and use the same method as we used for K-SVD$_P$ to find its optimal sparsity levels. We choose the SpaSM toolbox \cite{sjostrand2018spasm} based on piece-wise solution path ~\cite{Alpher19} to solve the LASSO problem in BPDN algorithm since it is faster than the popular glmnet package ~\cite{Alpher21}. 

\subsection{Reconstruction time}
For $\sigma =15,20,25$, we test the performance of three methods, BPDN, OMP and K-SVD$_P$, and run software in the Matlab$^\circledR$ R2017b environment on the Macbook with 2.9 GHz Intel$^\circledR$ Core$^\text{TM}$ i5 processor and 8G memory. For each noise level, we record the average reconstruction time among different images since the time expense is stable when images change. For $\sigma=50,75,100$, although BPDN may gain a bit higher quality, we need to abandon it since its time complexity is nearly 15 times that of the other two algorithms. This can be tell from Table 2. At the same time, we change to Matlab$^\circledR$ R2019a online environment which is faster to test the other two. The reconstruction time results are shown in Table 2. From the result, we can conclude that proposed K-SVD$_P$ framework is significant better than BPDN and is competitive to the OMP especially for images with high noise in terms of time.

\begin{table*}[!htbp]
\begin{center}
\resizebox{\textwidth}{40mm}{
\begin{tabular}{c|cc|cc|cc|cc|cc}
\hline
{Figure}& \multicolumn{2}{c|}{Map}& \multicolumn{2}{c|}{Man}& \multicolumn{2}{c|}{House}&\multicolumn{2}{c|}{Birdge}&\multicolumn{2}{c}{Lake}\\
\hline
{$\sigma/SSIM$}&OMP&K-SVD$_P$&OMP&K-SVD$_P$&OMP&K-SVD$_P$&OMP&K-SVD$_P$&OMP&K-SVD$_P$\\
\hline
{$15/0.829$}&\bf0.928&0.921&\bf0.910&0.907&\bf0.892&0.891&\bf0.931&0.923&\bf0.896&\bf0.896\\
{$20/0.753$}&\bf0.892&0.887&0.866&\bf0.867&0.846&\bf0.847&\bf0.898&0.893&0.846&\bf0.851\\
{$25/0.683$}&\bf0.854&0.848&0.819&\bf0.835&0.795&\bf0.816&\bf0.864&0.852&0.797&\bf0.828\\
{$50/0.443$}&0.678&\bf0.679&0.619&\bf0.707&0.596&\bf0.714&\bf0.683&0.677&0.596&\bf0.743\\
{$75/0.307$}&0.534&\bf0.599&0.475&\bf0.620&0.459&\bf0.626&0.536&\bf0.614&0.464&\bf0.649\\
{$100/0.223$}&0.430&\bf0.544&0.372&\bf0.556&0.360&\bf0.551&0.425&\bf0.557&0.370&\bf0.569\\
\hline
{Figure}& \multicolumn{2}{c|}{Airport}&\multicolumn{2}{c|}{Boat}&\multicolumn{2}{c|}{Airplane}&\multicolumn{2}{c|}{Lena}&\multicolumn{2}{c}{\bf{Average}}\\
\hline
{$\sigma/SSIM$}&OMP&K-SVD$_P$&OMP&K-SVD$_P$&OMP&K-SVD$_P$&OMP&K-SVD$_P$&OMP&K-SVD$_P$\\
\hline
{$15/0.829$}&\bf0.897&0.893&\bf0.897&0.895&0.874&\bf0.878&0.881&\bf0.889&\bf0.901&0.899\\
{$20/0.753$}&\bf0.850&0.848&\bf0.847&\bf0.847&0.818&\bf0.828&0.827&\bf0.839&0.854&\bf0.856\\
{$25/0.683$}&0.800&\bf0.810&0.795&\bf0.819&0.759&\bf0.800&0.769&\bf0.821&0.806&\bf0.825\\
{$50/0.443$}&0.585&\bf0.664&0.579&\bf0.711&0.544&\bf0.743&0.541&\bf0.754&0.602&\bf0.710\\
{$75/0.307$}&0.434&\bf0.574&0.432&\bf0.621&0.414&\bf0.635&0.395&\bf0.660&0.460&\bf0.622\\
{$100/0.223$}&0.333&\bf0.509&0.333&\bf0.540&0.331&\bf0.554&0.304&\bf0.575&0.362&\bf0.551\\
\hline
\end{tabular}}
\end{center}
\caption{Accuracy of the reconstruction in terms of the SSIM with $\sigma=15,20,25,50,75,100$ (The best are highlighted in bold)}
\end{table*}

\begin{figure*}[htbp]
\begin{center}
\begin{tabular}{ccccccccc}
\includegraphics[width=4.4cm]{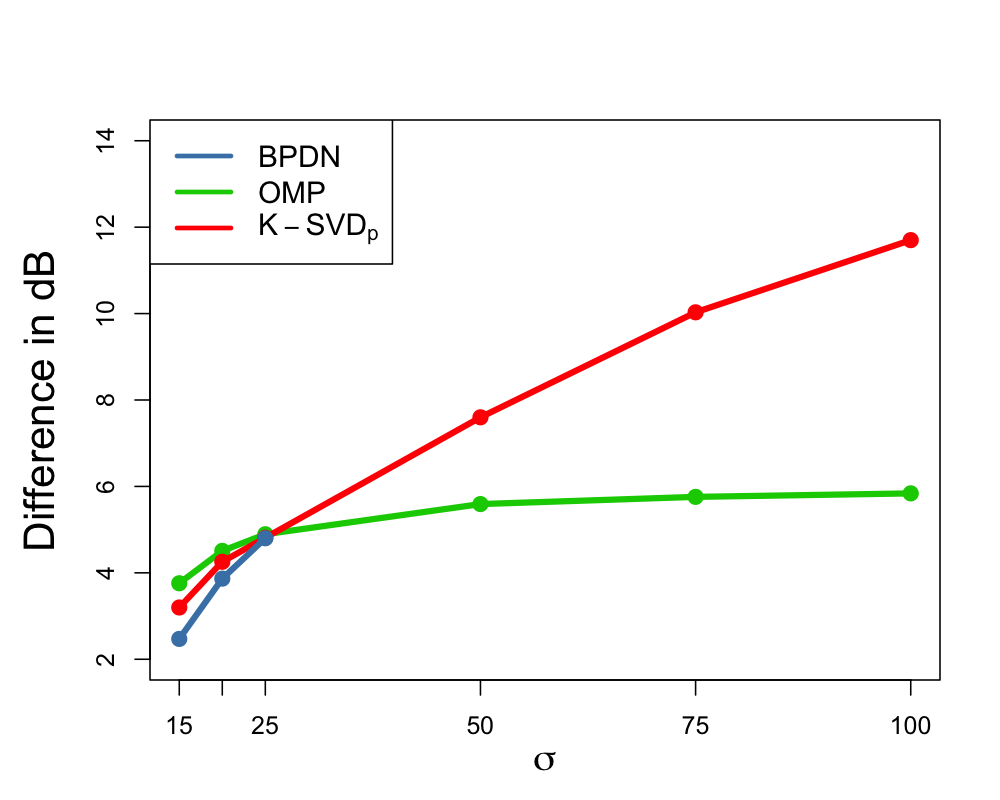}&
\includegraphics[width=4.4cm]{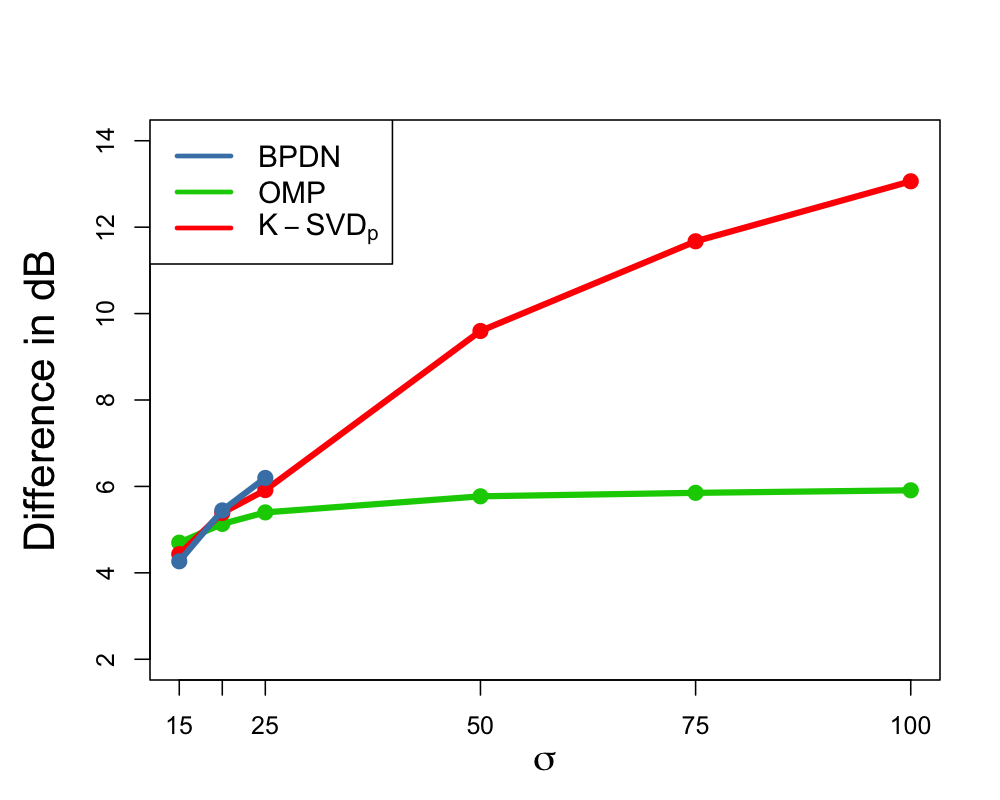}&
\includegraphics[width=4.4cm]{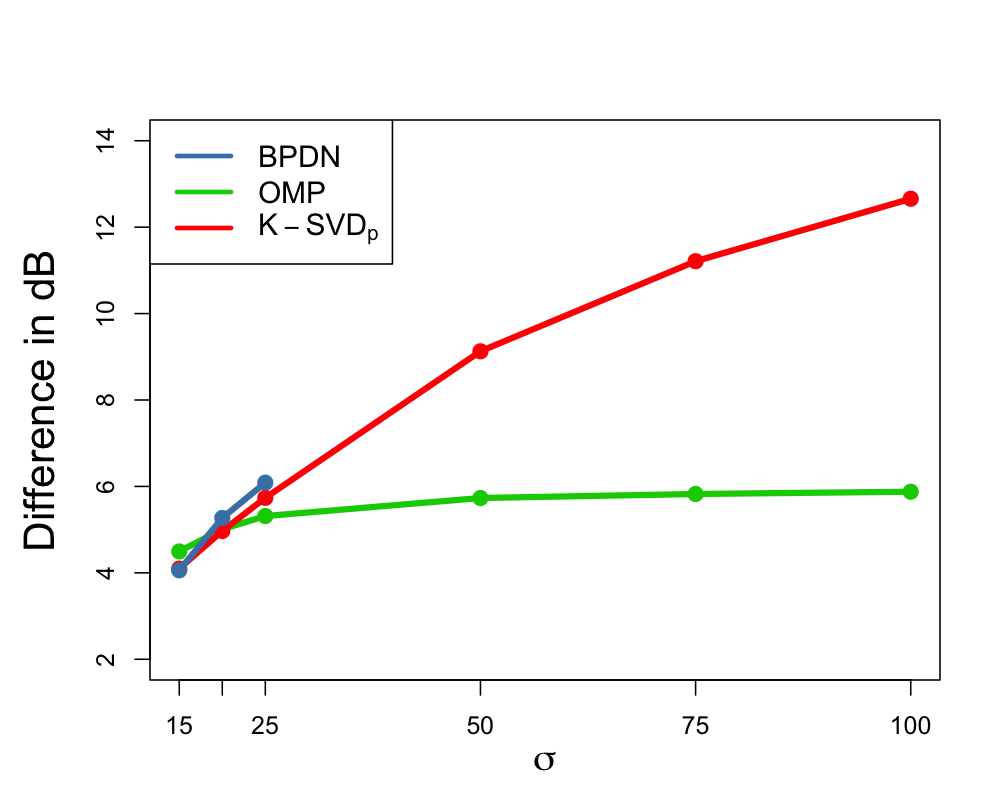}&\\
\footnotesize(a) Map&\footnotesize(b) Man&\footnotesize(c) House&\\
\includegraphics[width=4.4cm]{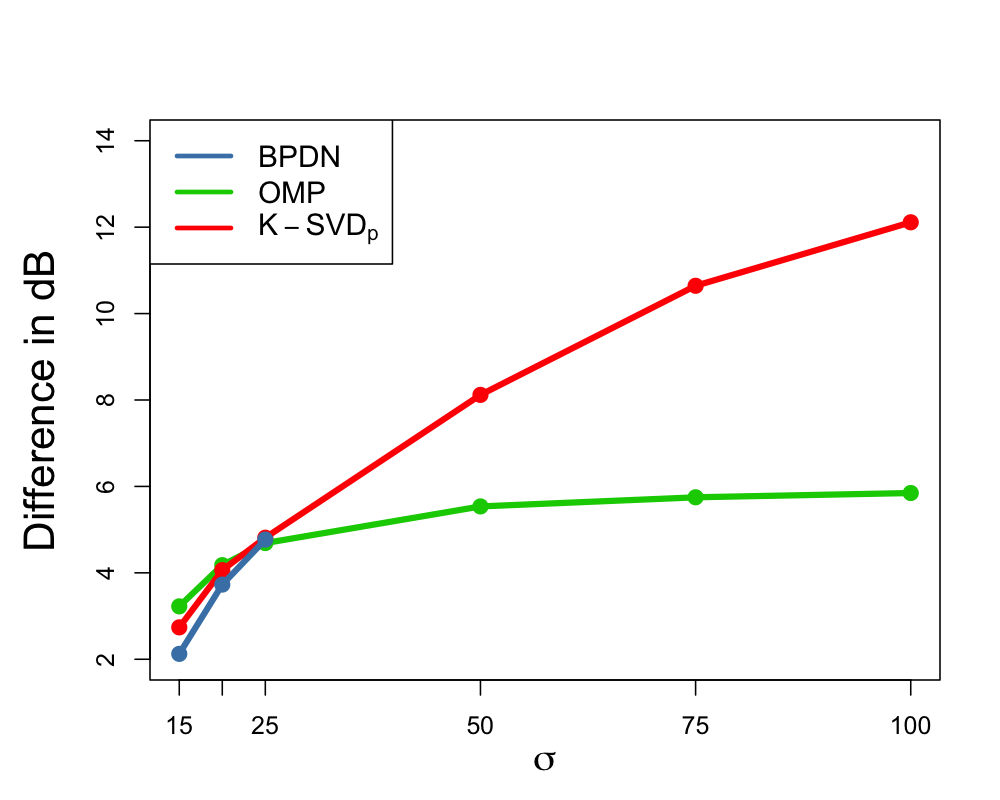}&
\includegraphics[width=4.4cm]{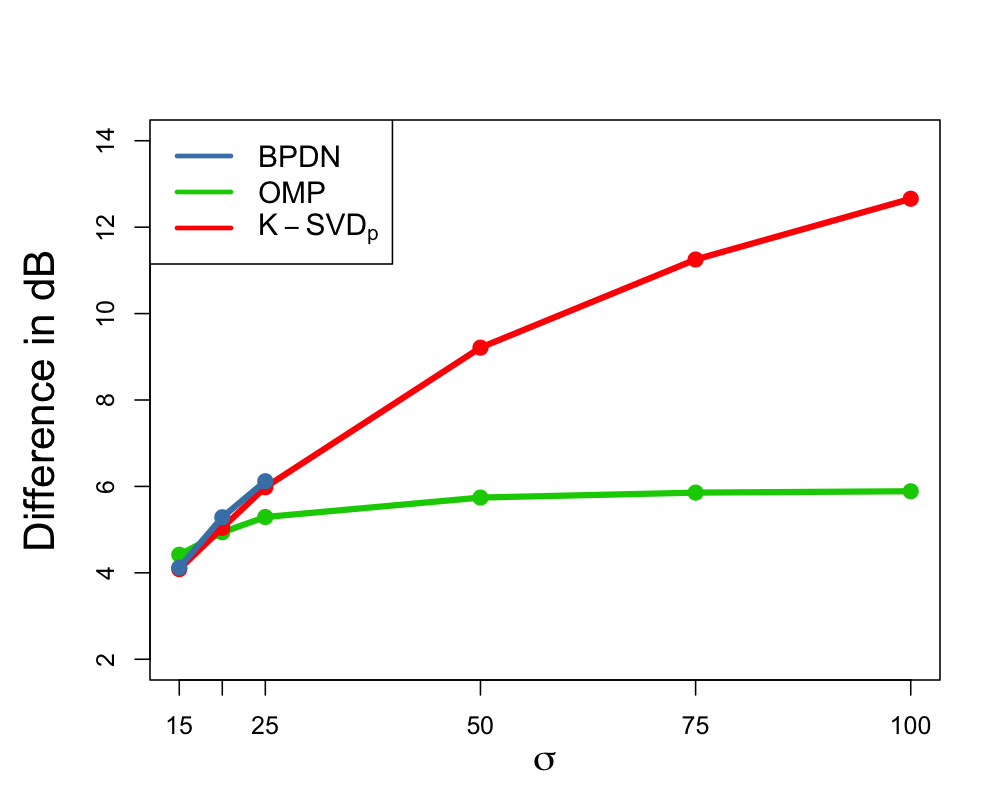}&
\includegraphics[width=4.4cm]{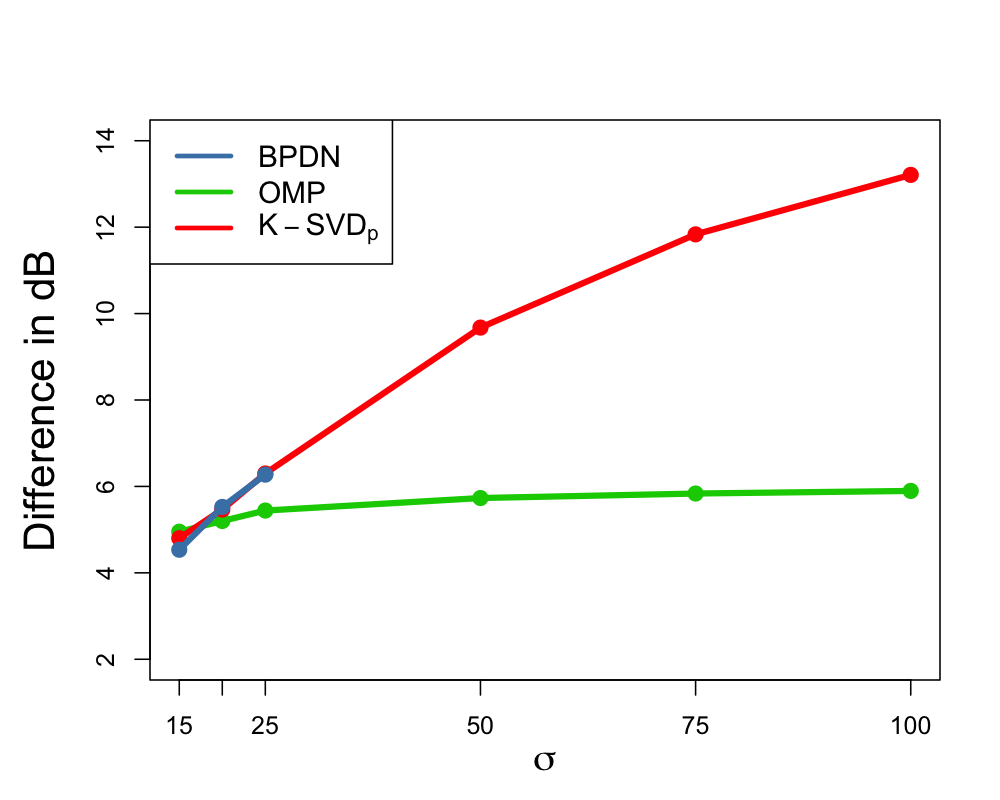}\\
\footnotesize(d) Bridge&\footnotesize(e) Lake&\footnotesize(f) Airport&\\
\includegraphics[width=4.4cm]{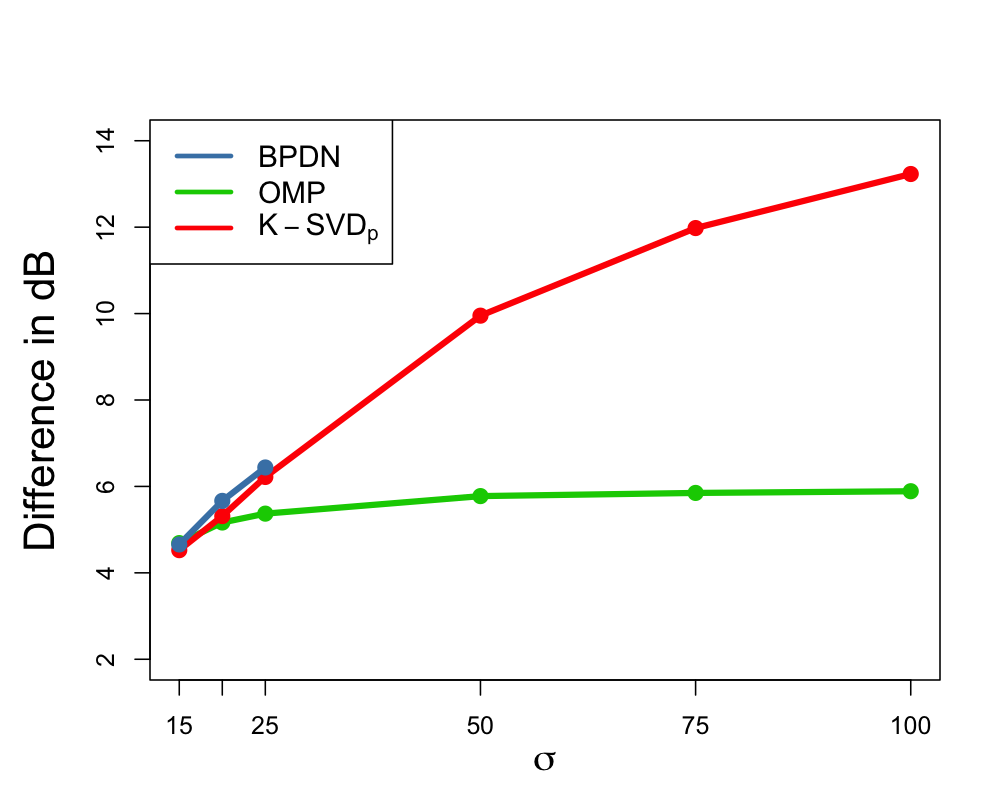}&
\includegraphics[width=4.4cm]{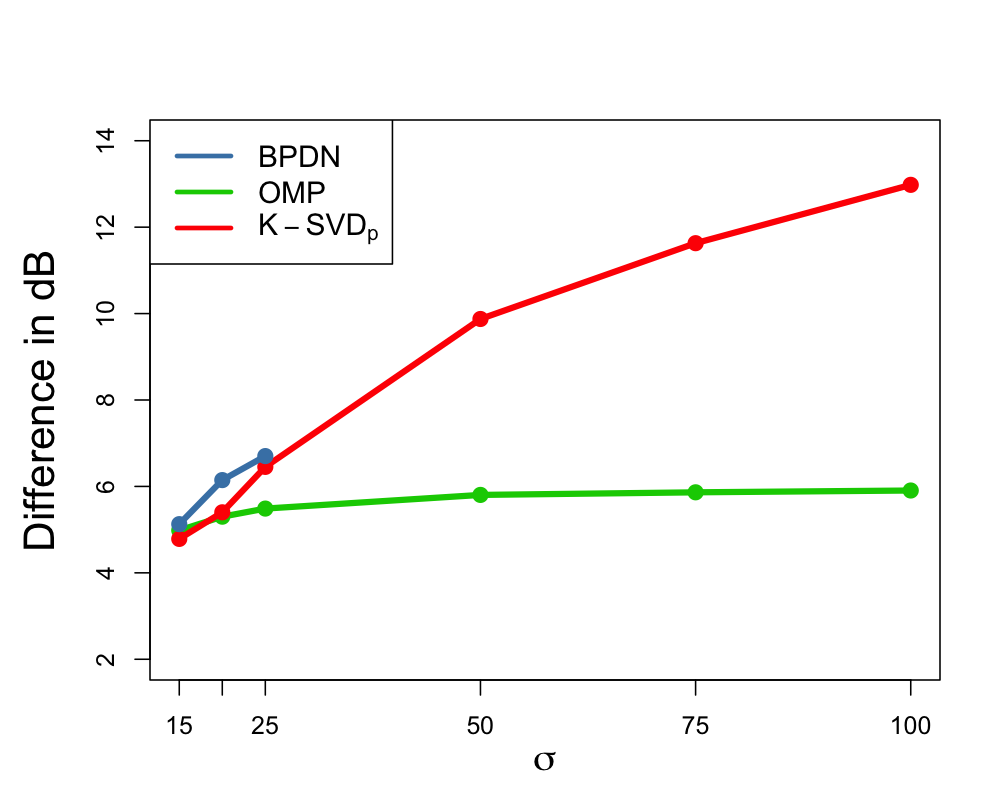}&
\includegraphics[width=4.4cm]{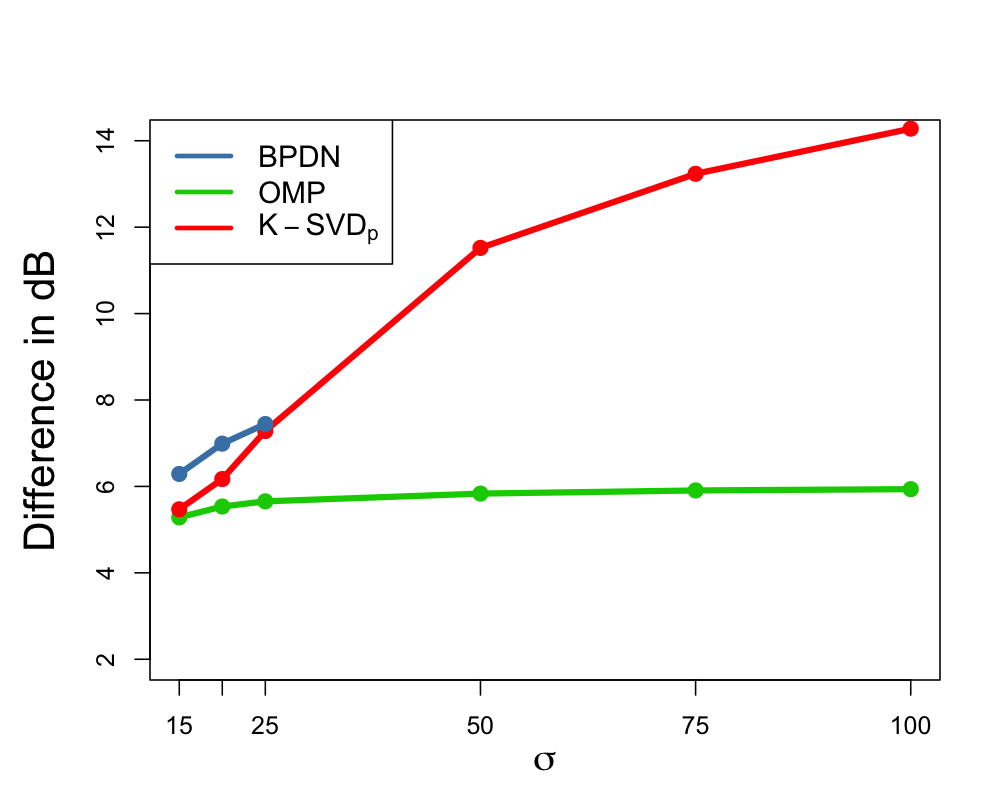}&\\
\footnotesize(g) Boat&\footnotesize(h) Airplane&\footnotesize(i) Lena&
\end{tabular}
\\
\caption{The difference of PSNR versus noise level of 9 images using different algorithms}
\label{fig:diff}
\end{center}
\end{figure*}

\subsection{PSNR comparison}
Table 3 shows PSNR results in low noisy cases. We can tell that the reconstruction performance of BPDN is better when the noise level is relatively high but the margin of difference with K-SVD$_P$ decreases when the noise level declines just as discussed in 3.2.2. The comparison between OMP and K-SVD$_P$ is exactly the opposite. Considering the time consumption and the fact that almost all the previous K-SVD benchmarks choose OMP as pursuit algorithm \cite{Claus06,abdelhamed2018high}, so the next comparisons are only between OMP and K-SVD$_P$. 

\begin{figure*}[htbp]
\begin{tabular}{cccc}
\includegraphics[width=4cm]{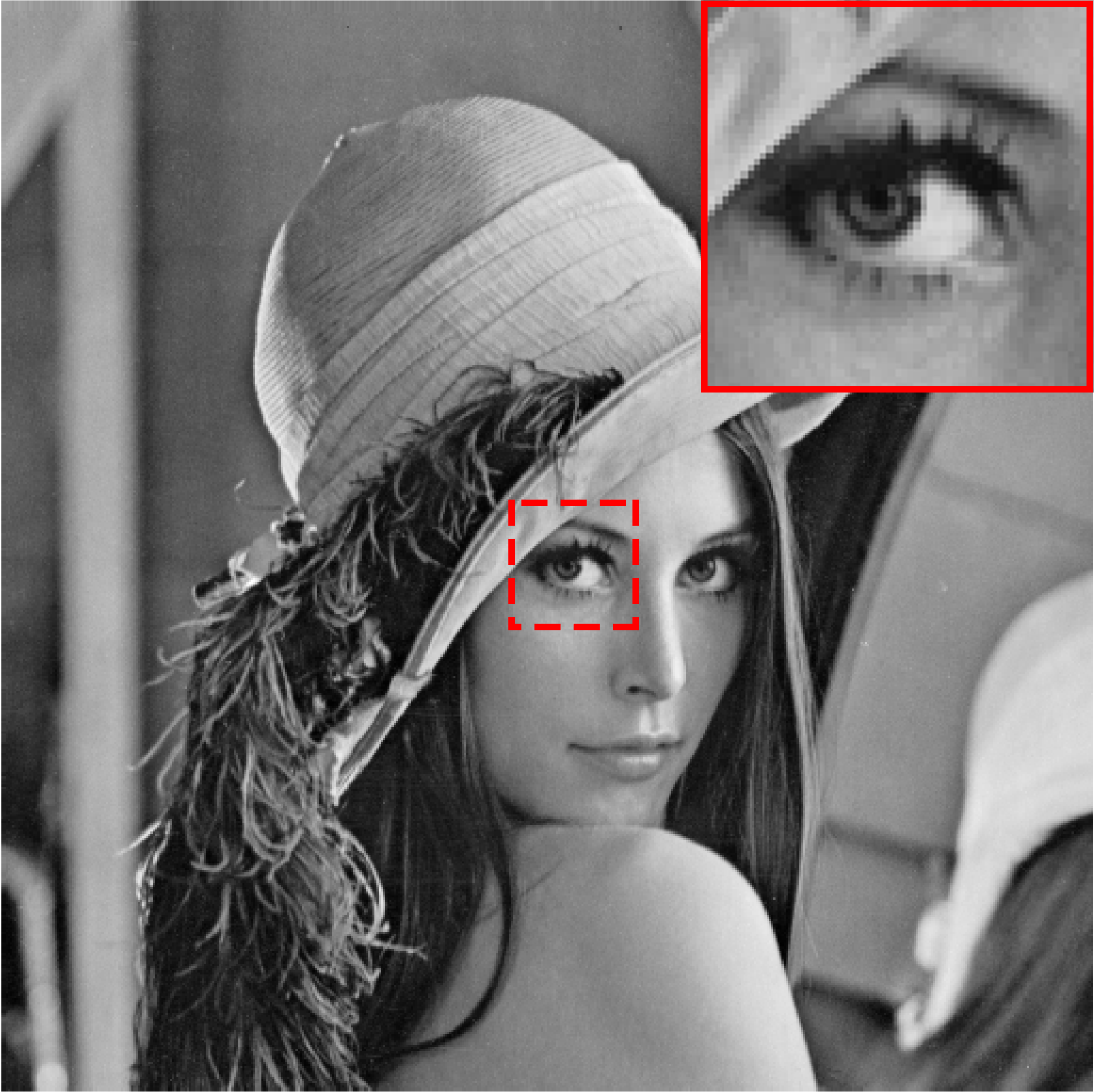}&
\includegraphics[width=4cm]{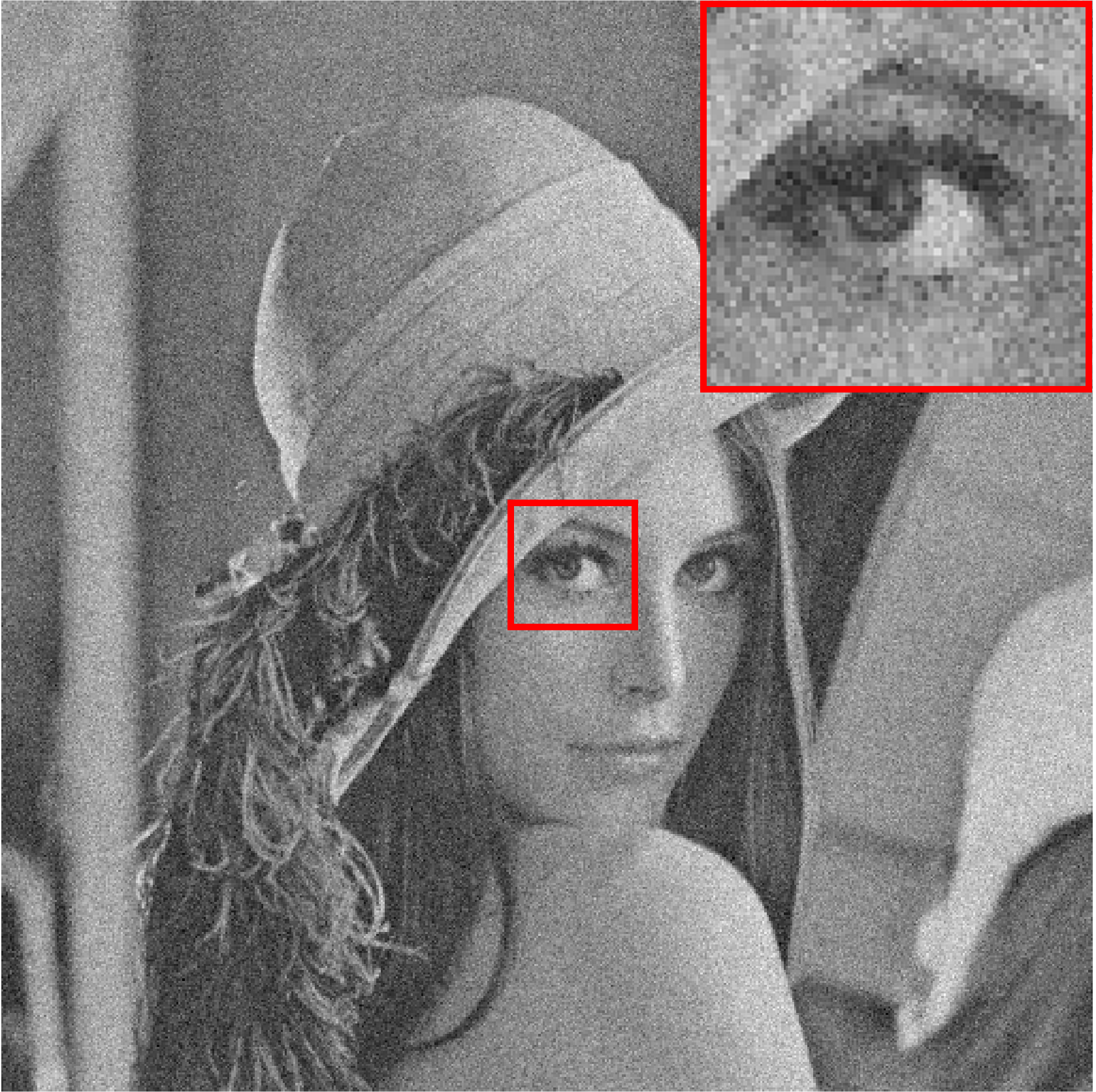}&
\includegraphics[width=4cm]{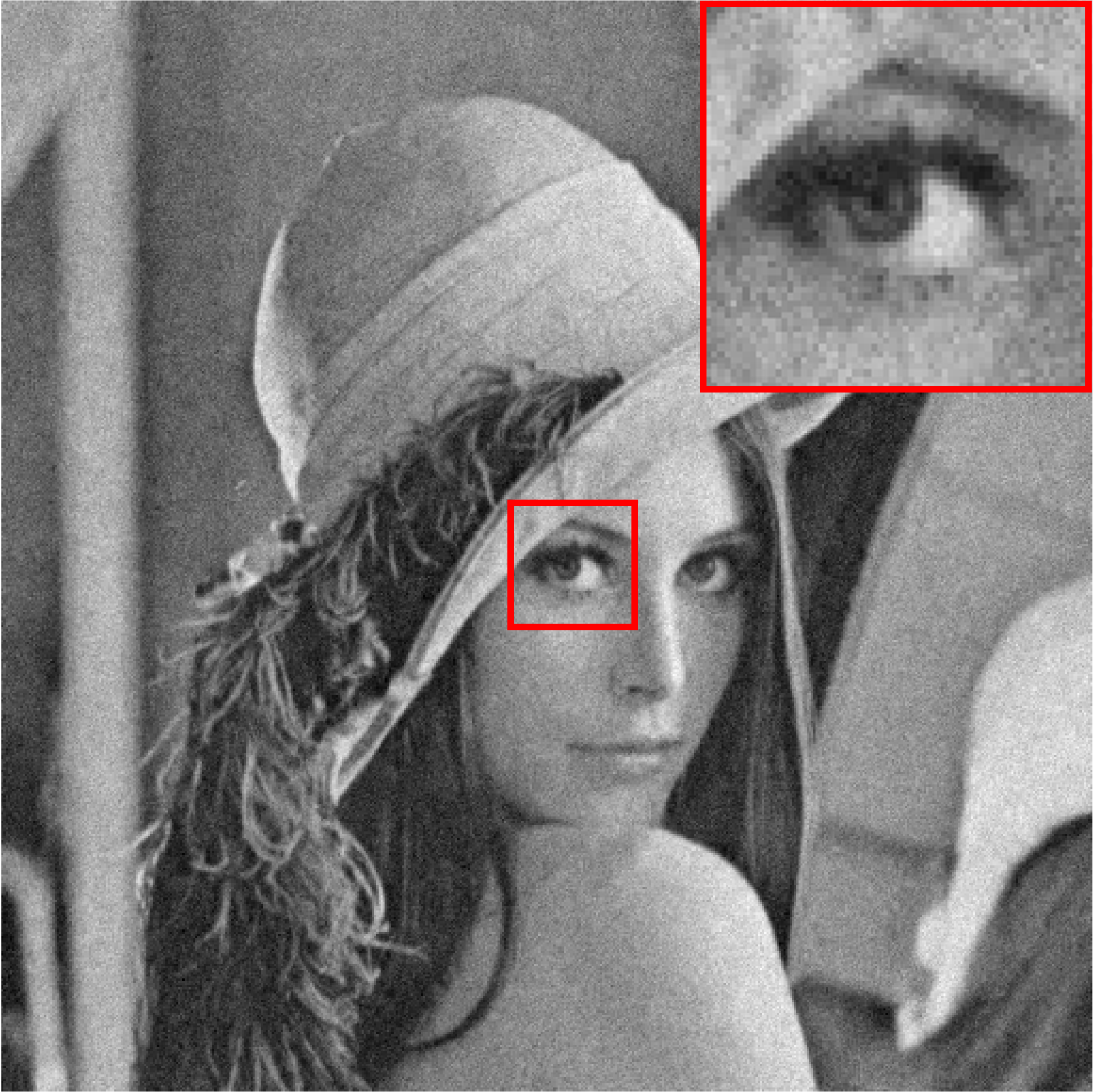}&
\includegraphics[width=4cm]{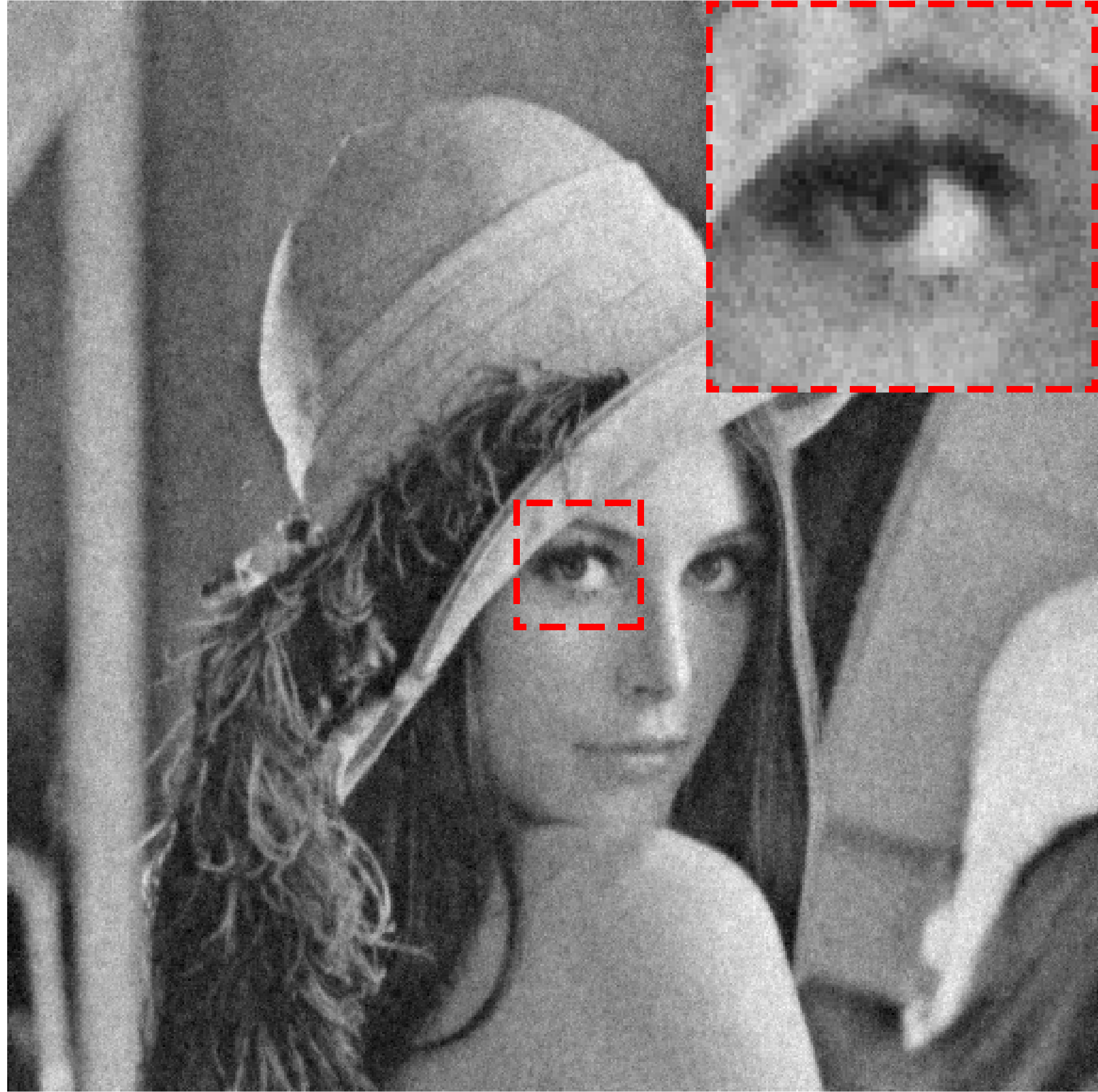}\\
\footnotesize(a) Original&\footnotesize(b) Noisy&\footnotesize(c) OMP (PSNR=27.64dB)&\footnotesize(d) K-SVD$_P$ (PSNR=28.27dB)
\end{tabular}
\\
\caption{Denoising results in Lenna, $\sigma=20$}
\label{fig:Lena}
\end{figure*}

\begin{figure*}
\begin{tabular}{cccc}
\includegraphics[width=4cm]{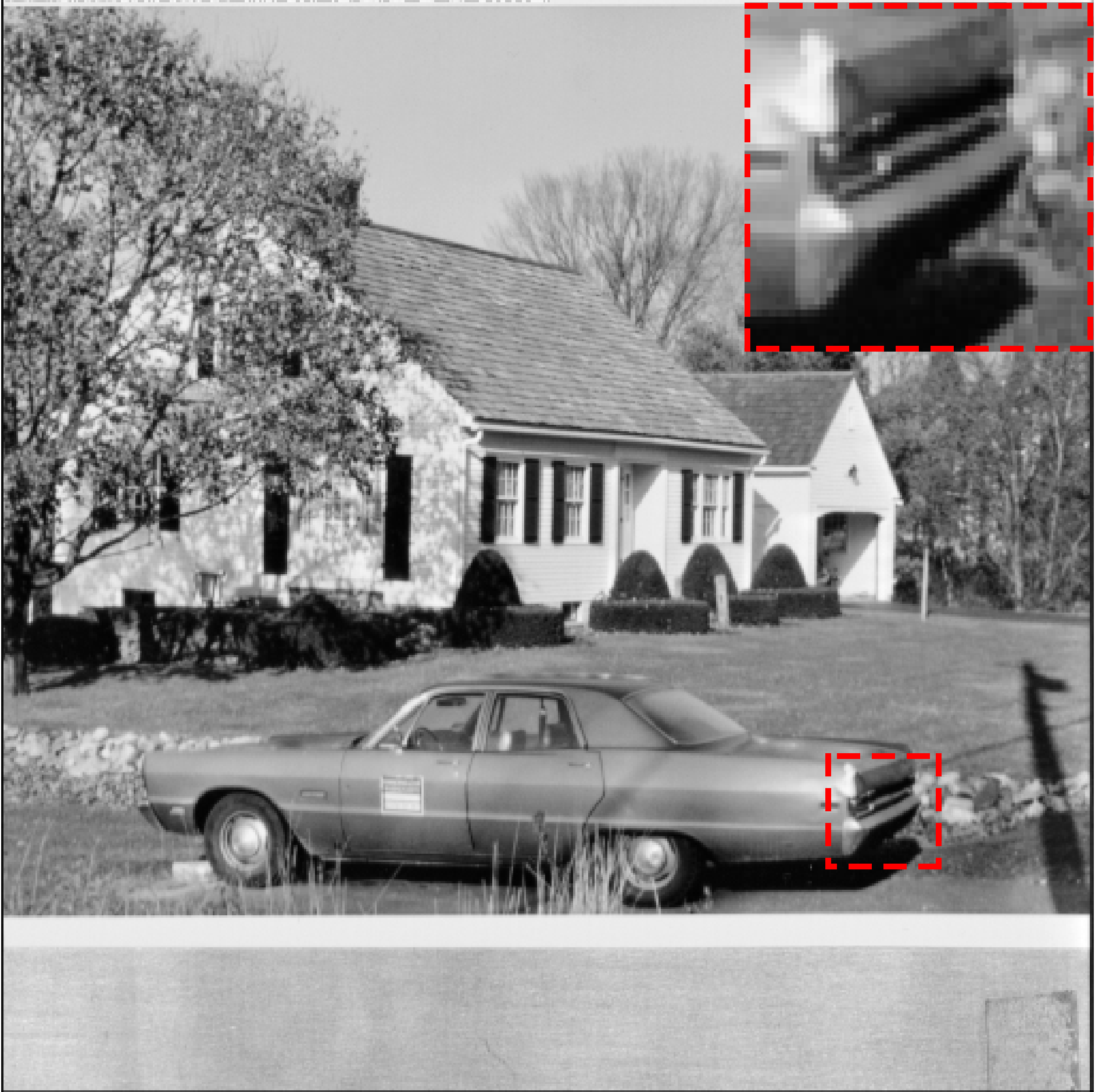}&
\includegraphics[width=4cm]{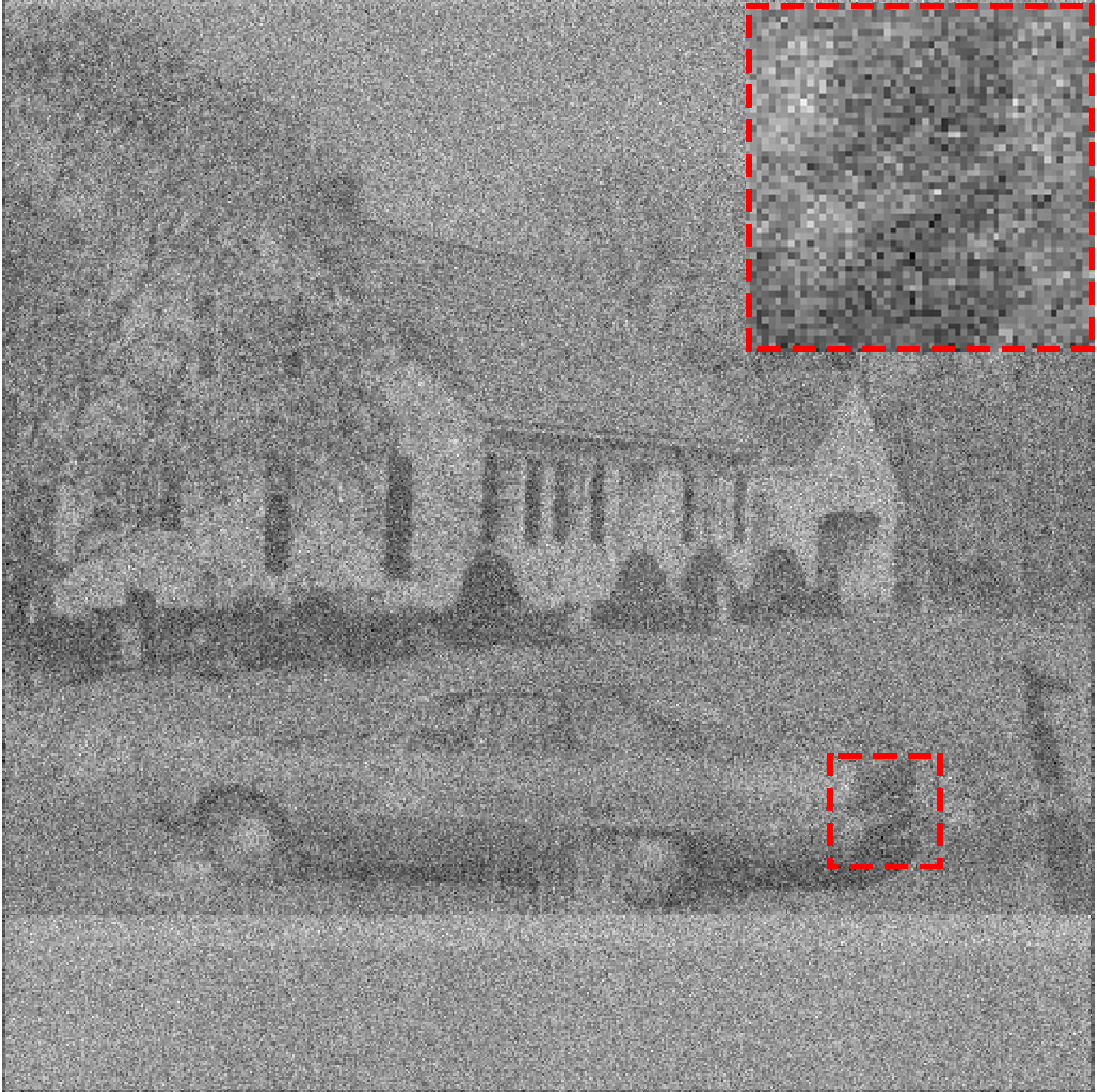}&
\includegraphics[width=4cm]{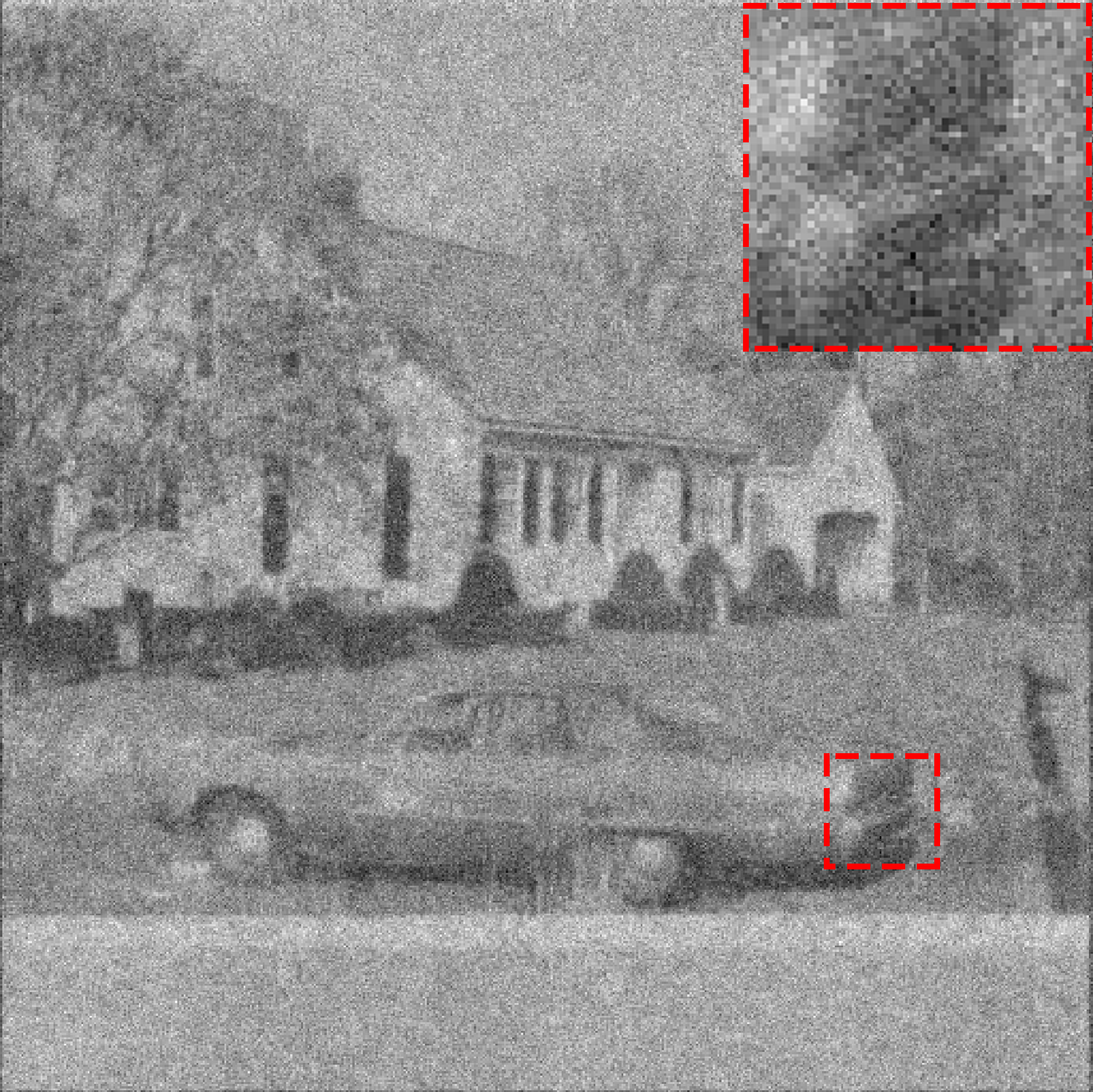}&
\includegraphics[width=4cm]{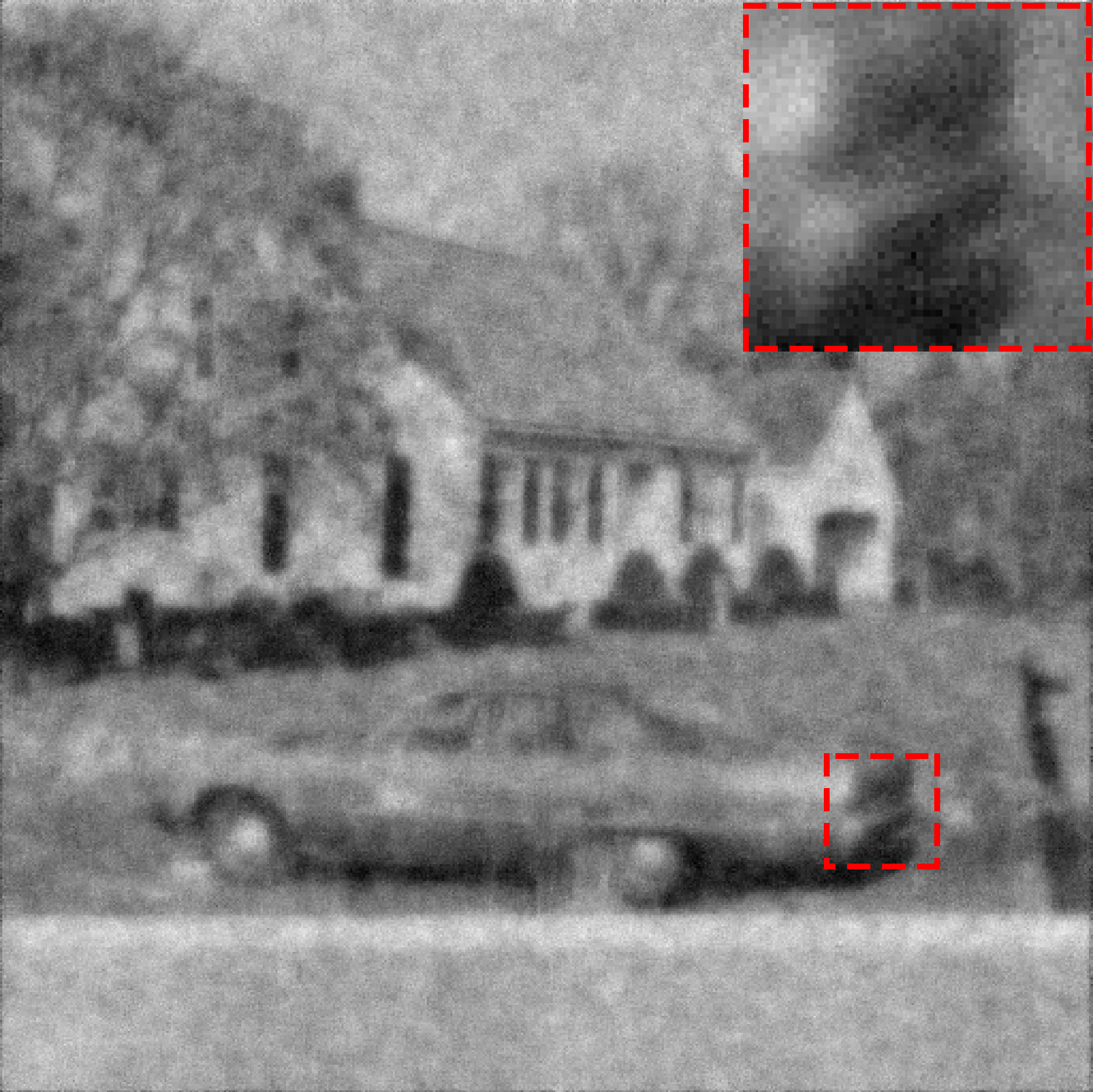}\\
\footnotesize(a) Original&\footnotesize(b) Noisy&\footnotesize(c) OMP (PSNR=16.46dB)&\footnotesize(d) K-SVD$_P$ (PSNR=21.85dB)
\end{tabular}
\\
\caption{Denoising results in House, $\sigma=75$}
\label{fig:House}
\end{figure*}

The high noise levels results of OMP and K-SVD$_P$ are shown in Table 4. For all images, K-SVD$_P$ outperforms OMP by a significant margin in high noise levels. In average, K-SVD$_P$ improves $3.68, 5.66, 6.99dB$ at $\sigma=50,75,100$ respectively. Figure 3 shows the difference of PSNR versus noise level for 9 images, and we can clearly see that K-SVD$_P$ is markedly potential as the noise level increases. This result can be expected. From 3.2.1, we know OMP will be inferior to K-SVD$_P$ at the same sparsity level. By preliminary experiments, we found the optimal $T_0$ for OMP is around 5. That's to say, once the optimal sparsity level of K-SVD$_P$ drops to lower than 5, it's impossible for OMP to defeat K-SVD$_P$. Then combined with the optimal sparsity level of K-SVD$_P$ showed in Table 1 and Figure 2, we can draw the conclusion.

\subsection{SSIM comparison}
Besides PSNR, structural similarity index (SSIM)\cite{wang2004image} is included to evaluate. Different from the PSNR, the SSIM is closer to the human visual effect since the correlation between image pixels is considered.
\begin{equation}
\operatorname{SSIM}(\mathbf{x}, \mathbf{y})=\frac{\left(2 \mu_{x} \mu_{y}+C_{1}\right)\left(2 \sigma_{x y}+C_{2}\right)}{\left(\mu_{x}^{2}+\mu_{y}^{2}+C_{1}\right)\left(\sigma_{x}^{2}+\sigma_{y}^{2}+C_{2}\right)}
\end{equation}
where $\mu_{x}$, $\mu_{y}$ are the mean intensity of the discrete signals, and $C1$, $C2$ are parameters to ensure the stability of SSIM. We use the default parameters and downsampling process.

Table 5 shows the SSIM of OMP and K-SVD$_P$ in $\sigma=15,20,25,50,75,100$. For images Man, House, Lake, Boat, Airplane and Lena which have clear objects in original images, K-SVD$_P$ is almost better than OMP at all noise levels. That is because a similar space in these images leads to high correlations between pixels. For those whose scenes are messy like Map, Bridge and Airport, results are similar to that in PSNR. PSNR is based on error sensitivity but the SSIM perceives image distortion by detecting whether the structural information changes. That's to say, although K-SVD$_P$ is slightly sensitive to error in low noise cases, it managed to maintain a spatial structure which is exactly where human vision is more concerned. 

\subsection{Visual comparison}
Figure 4 and Figure 5 show the denoising details for Lena with $\sigma=20$ and House with $\sigma=75$. To some extent, K-SVD$_P$ seems to employ a moving average of the image while OMP is more likely to operate on single points. So when looking at Lena's eye, the OMP processed one is more clear in single points such as eyeballs and eyeliners, while the K-SVD$_P$ operated one has less noise. The House results in $\sigma=75$ are more obvious. Only the K-SVD$_P$ restores the tail shape though the streaks on it are not clear enough. Figure 6 shows the final adaptive dictionaries trained by Man at $\sigma=50$. We can see the dictionary obtained by K-SVD$_P$ is highly structured compared to the OMP. 
\begin{figure*}[htbp]
\begin{center}
\begin{tabular}{cc}
\includegraphics[width=8cm]{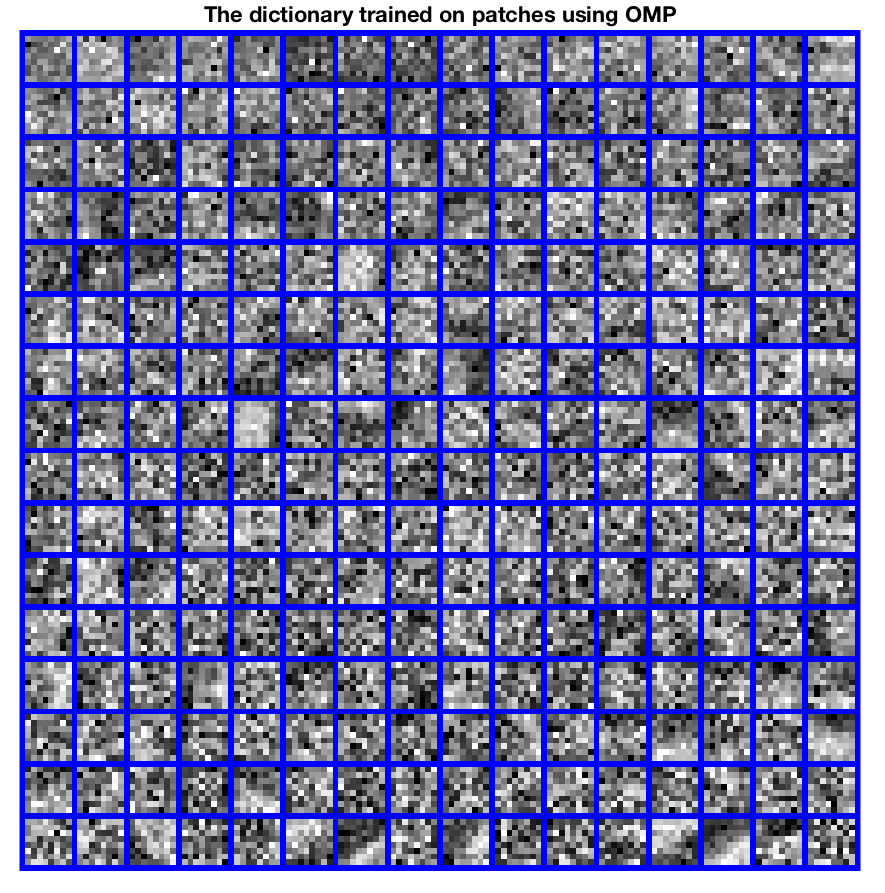}&
\includegraphics[width=8cm]{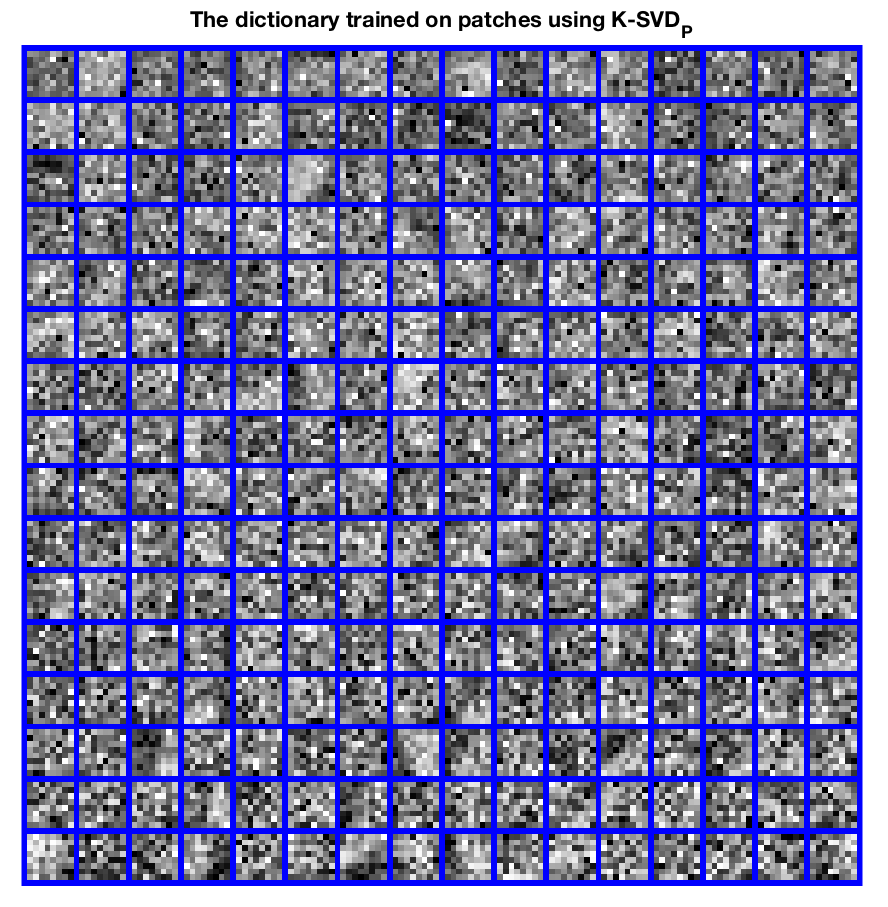}\\
\footnotesize(a) OMP&\footnotesize(b) K-SVD$_P$
\end{tabular}
\\
\caption{The trianed Dictionary of Man image, $\sigma=50$}
\label{fig:dic1}
\end{center}
\end{figure*}
\section{Conclusion and Future work}

In this paper, we proposed a new K-SVD$_P$ framework equipped with PDAS for sparse representation in image denoising. By introducing the primal-dual variables, the K-SVD$_P$ algorithm directly solves the best subset problem and presents a selection strategy that is different from the popular greedy algorithms and $\ell_{1}$ optimization. The explicit expression leads to low time complexity, while sufficient KKT condition leads to high accuracy, especially in high noisy cases. Moreover, the experiments demonstrate that the proposal is competitive and feasible compared with two state-of-the-art ways. 

The main benefits of our new K-SVD framework are:
\begin{itemize}
\item This new framework is superior to BPDN in time complexity and clarity at relatively low noise level;
\item In high noise cases, it achieves significantly better performance versus popular OMP algorithm and reduces the time complexity compared to BPDN which makes it possible to utilize;
\item Results of SSIM and visual comparisons reveal that it performs better on local patterns.
\end{itemize}
Future work will include a focus on improving the restoration performance of the K-SVD$_P$ framework at low noise levels and decreasing the time complexity further.

\section*{Acknowledgement}
This study is supported by NSFC (11801540), Natural Science Foundation of Guangdong (2017A030310572) and Fundamental Research Funds for the Central Universities \\(WK2040170015, WK2040000016).

{\small
\bibliographystyle{ieee}
\bibliography{egbib}

\begin{thebibliography}{10}\itemsep=-1pt

\bibitem{abdelhamed2018high}
A.~Abdelhamed, S.~Lin, and M.~S. Brown.
\newblock A high-quality denoising dataset for smartphone cameras.
\newblock In {\em Proceedings of the IEEE Conference on Computer Vision and
  Pattern Recognition}, pages 1692--1700, 2018.

\bibitem{Alpher07}
M.~Aharon, M.~Elad, A.~Bruckstein, et~al.
\newblock {K-SVD}: An algorithm for designing overcomplete dictionaries for
  sparse representation.
\newblock {\em IEEE Transactions on signal processing}, 54(11):4311, 2006.

\bibitem{baloch2017image}
G.~Baloch and H.~Ozkaramanli.
\newblock Image denoising via correlation-based sparse representation.
\newblock {\em Signal, Image and Video Processing}, 11(8):1501--1508, 2017.

\bibitem{bergeaud1995matching}
F.~Bergeaud and S.~Mallat.
\newblock Matching pursuit of images.
\newblock In {\em Proceedings., International Conference on Image Processing},
  volume~1, pages 53--56. IEEE, 1995.

\bibitem{chen2001atomic}
S.~S. Chen, D.~L. Donoho, and M.~A. Saunders.
\newblock Atomic decomposition by basis pursuit.
\newblock {\em SIAM review}, 43(1):129--159, 2001.

\bibitem{Alpher30}
Y.~Chen.
\newblock Fast dictionary learning for noise attenuation of multidimensional
  seismic data.
\newblock {\em Geophysical Journal International}, 209(1):21--31, 2017.

\bibitem{Alpher25}
W.~Dong, L.~Zhang, G.~Shi, and X.~Li.
\newblock Nonlocally centralized sparse representation for image restoration.
\newblock {\em IEEE transactions on Image Processing}, 22(4):1620--1630, 2013.

\bibitem{Alpher18}
B.~Efron, T.~Hastie, I.~Johnstone, R.~Tibshirani, et~al.
\newblock Least angle regression.
\newblock {\em The Annals of statistics}, 32(2):407--499, 2004.

\bibitem{Alpher23}
M.~Elad and M.~Aharon.
\newblock Image denoising via sparse and redundant representations over learned
  dictionaries.
\newblock {\em IEEE Transactions on Image processing}, 15(12):3736--3745, 2006.

\bibitem{Alpher31}
Q.~Guo, C.~Zhang, Y.~Zhang, and H.~Liu.
\newblock An efficient {SVD-based} method for image denoising.
\newblock {\em IEEE transactions on Circuits and Systems for Video Technology},
  26(5):868--880, 2016.

\bibitem{hastie2017extended}
T.~Hastie, R.~Tibshirani, and R.~J. Tibshirani.
\newblock Extended comparisons of best subset selection, forward stepwise
  selection, and the lasso.
\newblock {\em arXiv preprint arXiv:1707.08692}, 2017.

\bibitem{Alpher35}
K.~Ito and K.~Kunisch.
\newblock A variational approach to sparsity optimization based on {Lagrange}
  multiplier theory.
\newblock {\em Inverse problems}, 30(1):015001, 2013.

\bibitem{jhang2016high}
J.-W. Jhang and Y.-H. Huang.
\newblock A {high-SNR} projection-based atom selection {OMP} processor for
  compressive sensing.
\newblock {\em IEEE Transactions on Very Large Scale Integration (VLSI)
  Systems}, 24(12):3477--3488, 2016.

\bibitem{liu2019mixed}
Y.~Liu, S.~Canu, P.~Honeine, and S.~Ruan.
\newblock Mixed {Integer} {Programming} for {Sparse} {Coding}: {Application} to
  {Image} {Denoising}.
\newblock {\em IEEE Transactions on Computational Imaging}, 2019.

\bibitem{Alpher29}
X.~Lu and X.~L{\"u}.
\newblock {ADMM} for image restoration based on nonlocal simultaneous sparse
  {Bayesian} coding.
\newblock {\em Signal Processing: Image Communication}, 70:157--173, 2019.

\bibitem{Alpher37}
B.~K. Natarajan.
\newblock Sparse approximate solutions to linear systems.
\newblock {\em SIAM journal on computing}, 24(2):227--234, 1995.

\bibitem{olshausen1997sparse}
B.~A. Olshausen and D.~J. Field.
\newblock Sparse coding with an overcomplete basis set: A strategy employed by
  v1?
\newblock {\em Vision research}, 37(23):3311--3325, 1997.

\bibitem{Claus06}
T.~Plotz and S.~Roth.
\newblock Benchmarking denoising algorithms with real photographs.
\newblock In {\em Proceedings of the IEEE Conference on Computer Vision and
  Pattern Recognition}, pages 1586--1595, 2017.

\bibitem{Alpher33}
M.~Protter and M.~Elad.
\newblock Image sequence denoising via sparse and redundant representations.
\newblock {\em IEEE Transactions on Image Processing}, 18(1):27--35, 2009.

\bibitem{Alpher21}
J.~Qian, T.~Hastie, J.~Friedman, R.~Tibshirani, and N.~Simon.
\newblock Glmnet for matlab.
\newblock {\em Accessed: Nov}, 13:2017--2017, 2013.

\bibitem{Alpher26}
M.~E. R.~Rubinstein, A. M.~Bruckstein.
\newblock Dictionaries for sparse representation modeling.
\newblock {\em IEEE Proc.}, 98(6):1045--1057, June 2010.

\bibitem{Alpher19}
S.~Rosset and J.~Zhu.
\newblock Piecewise linear regularized solution paths.
\newblock {\em The Annals of Statistics}, pages 1012--1030, 2007.

\bibitem{Alpher27}
R.~Rubinstein, A.~M. Bruckstein, and M.~Elad.
\newblock Dictionaries for sparse representation modeling.
\newblock {\em Proceedings of the IEEE}, 98(6):1045--1057, 2010.

\bibitem{sajjad2016basis}
M.~Sajjad, I.~Mehmood, N.~Abbas, and S.~W. Baik.
\newblock Basis pursuit denoising-based image superresolution using a redundant
  set of atoms.
\newblock {\em Signal, Image and Video Processing}, 10(1):181--188, 2016.

\bibitem{sjostrand2018spasm}
K.~Sj{\"o}strand, L.~H. Clemmensen, R.~Larsen, G.~Einarsson, and B.~K.
  Ersb{\o}ll.
\newblock Spasm: A matlab toolbox for sparse statistical modeling.
\newblock {\em Journal of Statistical Software}, 84(10), 2018.

\bibitem{Claus05}
J.~Sulam, B.~Ophir, and M.~Elad.
\newblock Image denoising through multi-scale learnt dictionaries.
\newblock In {\em 2014 IEEE International Conference on Image Processing
  (ICIP)}, pages 808--812. IEEE, 2014.

\bibitem{Alpher15}
R.~Tibshirani.
\newblock Regression shrinkage and selection via the lasso.
\newblock {\em Journal of the Royal Statistical Society: Series B
  (Methodological)}, 58(1):267--288, 1996.

\bibitem{tropp2004greed}
J.~A. Tropp.
\newblock Greed is good: Algorithmic results for sparse approximation.
\newblock {\em IEEE Transactions on Information theory}, 50(10):2231--2242,
  2004.

\bibitem{tropp2007signal}
J.~A. Tropp and A.~C. Gilbert.
\newblock Signal recovery from random measurements via orthogonal matching
  pursuit.
\newblock {\em IEEE Transactions on information theory}, 53(12):4655--4666,
  2007.

\bibitem{wang2004image}
Z.~Wang, A.~C. Bovik, H.~R. Sheikh, E.~P. Simoncelli, et~al.
\newblock Image quality assessment: from error visibility to structural
  similarity.
\newblock {\em IEEE transactions on image processing}, 13(4):600--612, 2004.

\bibitem{aawen2017bess}
C.~Wen, A.~Zhang, S.~Quan, and X.~Wang.
\newblock Bess: An {R} {Package} for {Best} {Subset} {Selection} in {Linear},
  {Logistic} and {CoxPH} {Models}.
\newblock {\em arXiv preprint arXiv:1709.06254}, 2017.

\bibitem{Alpher20}
H.~Zou.
\newblock The adaptive lasso and its oracle properties.
\newblock {\em Journal of the American statistical association},
  101(476):1418--1429, 2006.

\end{thebibliography}
}

\end{document}